\documentclass[9.5pt,twocolumn,twoside]{IEEEtran}
\pdfoutput=1
\usepackage{graphicx}
\usepackage{amsmath,amsfonts,amssymb,amscd,amsthm,xspace}
\usepackage{import}
\usepackage{xcolor}
\usepackage{array}
\usepackage{ragged2e}
\usepackage{subfig}
\usepackage{mathtools}
\usepackage{math}
\usepackage{multirow}
\usepackage{enumitem}
\usepackage[font=small,labelfont=rm,textfont=rm]{caption}
\usepackage{multirow}
\usepackage{booktabs}
\usepackage[lined,boxed,commentsnumbered,ruled,norelsize]{algorithm2e}
\makeatletter
\newcommand{\removelatexerror}{\let\@latex@error\@gobble}
\makeatother
\usepackage{balance}
\usepackage[noadjust,sort]{cite} 
\usepackage{url}
\usepackage{tikz}
\usepackage{relsize}
\usetikzlibrary{positioning}
\tikzset{fontscale/.style = {font=\relsize{#1}}}
\usepackage{pbox}

\usepackage[final]{review}
\setcoverletter{coverletter-R2.tex}
\setrevision{2}

\newcommand{\eg}{e.g. }

\title{Variational Bayesian Inference for Audio-Visual Tracking of Multiple Speakers}
\author{Yutong Ban, Xavier Alameda-Pineda, 
Laurent Girin and Radu Horaud
\IEEEcompsocitemizethanks{
Y. Ban, X. Alameda-Pineda and R. Horaud are with Inria Grenoble Rh\^one-Alpes, Montbonnot Saint-Martin, France. L. Girin is with GIPSA Lab, Univ. Grenoble 
Alpes, France. 
}
}

\definecolor{bleudefrance}{rgb}{0.19, 0.55, 0.91}
\definecolor{babyblue}{rgb}{0.54, 0.81, 0.94}
\definecolor{ashgrey}{rgb}{0.7, 0.75, 0.71}
\definecolor{awesome}{rgb}{1.0, 0.13, 0.32}
\definecolor{bittersweet}{rgb}{1.0, 0.44, 0.37}
\definecolor{celadon}{rgb}{0.67, 0.88, 0.69}
\definecolor{emerald}{rgb}{0.31, 0.78, 0.47}

\begin{document}
\maketitle
\begin{abstract}
In this paper we address the problem of tracking multiple speakers via the fusion of visual and auditory information. We propose to exploit the complementary nature and roles of these two modalities in order to accurately estimate smooth trajectories of the tracked persons, to deal with the partial or total absence of one of the modalities over short periods of time, and to estimate the acoustic status -- either speaking or silent -- of each tracked person over time. We propose to cast the problem at hand into a generative audio-visual fusion (or association) model formulated as a latent-variable temporal graphical model. This may well be viewed as the problem of maximizing the posterior joint distribution of a set of continuous and discrete latent variables given the past and current observations, which is intractable. We propose a variational inference model which amounts approximating the joint distribution with a factorized distribution. The solution takes the form of a closed-form expectation maximization procedure. We describe in detail the inference algorithm, we evaluate its performance and we compare it with several baseline methods. These experiments show that the proposed audio-visual tracker performs well in informal meetings involving a time-varying number of people.

\end{abstract}

\section{Introduction}
\label{sec:introduction}
In this paper we address the problem of tracking multiple speakers via the fusion of visual and auditory information \cite{gatica2007audiovisual,hospedales08structure,NavqiMiaoChambers2010,kilicc2015audio,schult2015information,barnard2016mean,kilicc2016mean}. We propose to exploit the complementary nature of these two modalities in order to accurately estimate the position of each person at each time step, to deal with the partial or total absence of one of the modalities over short periods of time, and to estimate the acoustic status, either speaking or silent, of each tracked person. We propose to cast the problem at hand into a generative audio-visual fusion (or association) model formulated as a latent-variable temporal graphical model. We propose a tractable solver via a variational approximation. 

\begin{figure}[t!]
\begin{center}
\includegraphics[width = 0.45\textwidth]{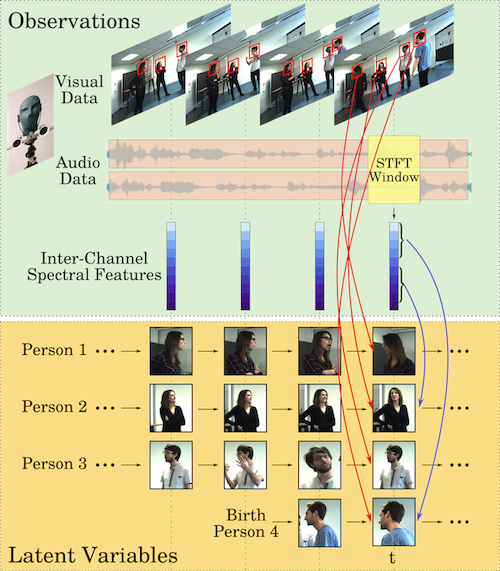}
\end{center}
\caption{\label{fig:overview} Multiple speaker tracking is cast into the framework of Bayesian inference. Visual observations (person detections) and audio 
observations (inter-channel spectral features) are assigned to  continuous latent variables (i.e. speaker positions) via discrete latent variables (one for 
each observation). As shown, the algorithm is causal (it uses only past and present observations) and incorporates a birth process to account for not yet 
seen/heard persons.
}
\end{figure}

We are particularly interested in tracking people involved in informal meetings and social gatherings, e.g. Fig.~\ref{fig:overview}. In this type of scenarios, 
participants wander around, cross each other, move in and out the camera field of view, take speech turns, etc. Acoustic room conditions, e.g. reverberation, 
and overlapping audio sources of various kinds drastically deteriorate or modify the microphone signals. Likewise, occluded persons, lighting conditions and  
mid-range camera distance complicate the task of visual processing. It is therefore impossible to gather reliable and continuous flows of visual \textbf{and} 
audio observations. Hence one must design a fusion and tracking method that is able to deal with intermittent visual and audio data.

We propose a multi-speaker tracking method based on a dynamic Bayesian model that fuses audio and visual information over time from their respective observation 
spaces. This may well be viewed as a generalization of single-observation and single-target Kalman filtering -- which yields an exact recursive solution -- to 
multiple observations and multiple targets, which makes the exact recursive solution computationally intractable. We propose a variational approximation of the 
joint posterior distribution over the continuous variables (positions and velocities of tracked persons) and discrete variables  (observation-to-person 
associations) at each time step, given all the past and present audio and visual observations. The proposed approximation consists on factorizing the joint 
distribution. We obtain a variational expectation maximisation (VEM) algorithm that is not only computationally tractable, but also very efficient.

In general, multiple object tracking consists of the temporal estimation of the kinematic state of each object, i.e. position and velocity. In computer vision,  
local descriptors are used to better discriminate between objects, e.g. person detectors/descriptors based on hand-crafted features \cite{ba2016line} or on deep 
neural networks \cite{BaeYoon2018}. If the tracked objects emit sounds, their states can be inferred as well using sound-source localization techniques combined 
with tracking, e.g. \cite{valin2007robust}. These techniques are often based on the estimation of the sound's direction of arrival (DOA) using a microphone array, e.g. 
\cite{lombard2011tdoa}, or on a steered beamformer \cite{valin2007robust}. DOA estimation can be carried out either in the temporal domain \cite{alameda2014geometric}, or in the spectral (Fourier) domain 
\cite{dorfan2015tree}. However, spectral-domain DOA estimation methods are more robust than temporal-domain methods, in particular in the presence of background 
noise and reverberation \cite{li2016estimation,li2017multiple}. \addnote[steeredbeamer]{2}{The multiple sound-source localization and tracking method of \cite{valin2007robust} combines a steered beamformer with a particle filter. The loudest sound source is detected first, the second loudest one is next detected, etc., and up to four sources. This leads to many false detections. Particle filtering is combined with source-to-track assignment  probabilities in order to determine whether a newly detected source is a false detection, a source that is currently being tracked, or a new source. In practice, this method requires several empirically defined thresholds.}

Via proper camera-microphone calibration, audio and visual observations can be aligned such that a DOA corresponds to a 2D location in the image plane. In this paper we adopt the audio-visual alignment method of \cite{deleforge2015colocalization}, which learns a mapping from the space spanned by \textit{inter-channel spectral features} (audio features) to the space of source locations, which in our case corresponds to the image plane. Interestingly, the method of  \cite{deleforge2015colocalization} estimates both this mapping and its inverse via a closed-form EM algorithm. Moreover, this allows us to exploit the richness of representing acoustic signals in the short-time Fourier domain \cite{gold2011speech} and to extract noise- and reverberation-free audio features \cite{li2016estimation}. 

We propose to represent the audio-visual fusion problem via two sets of independent variables, i.e. visual-feature-to-person and audio-feature-to-person sets of assignment variables. An interesting characteristic of this way of doing is that the proposed tracking algorithm can choose to use visual features, audio features, or a combination of both, and this choice can be made independently for every person and for every time step. 
Indeed, audio and visual information are rarely available simultaneously and continuously. Visual information suffers from limited camera field-of-view, occlusions, false positives, missed detections, etc. Audio information is often corrupted by room acoustics, environmental noise and overlapping acoustic signals. In particular speech signals are sparse, non-stationary and are emitted intermittently, with silence intervals between speech utterances. Hence a robust audio-visual tracking must explicitly take into account the temporal sparsity of the two modalities and this is exactly what is proposed in this paper.

\addnote[av163-ospa]{1}{
We use the AV16.3 \cite{lathoud2004av16} and the AVDIAR \cite{gebru2018a} datasets to evaluate the performance of the proposed audio-visual tracker. We use the Multiple Object Tracking (MOT) metrics and the Optimal Sub Pattern Assignment fo Tracks (OSPA-T) metrics to quantitatively assess method performance. 
MOT and in particular MOTA (tracking accuracy), which combines false positives, false negatives, identity switches, by comparing the estimated tracks with the 
ground-truth trajectories, is a commonly used score to assess the quality of a multiple person tracker.\footnote{\url{https://motchallenge.net/}} OSPA-T 
measures the distance between two point sets and hence it is also useful to compare ground-truth tracks with estimated tracks in the context of multi-target 
tracking \cite{ristic2011metric}.
We use MOT and OSPA-T metrics to compare our method with two recently proposed audio-visual tracking methods \cite{kilicc2015audio,kilicc2016mean} and with a visual tracker \cite{ba2016line}. An interesting outcome of the proposed method is that speaker diarization, i.e. who speaks when, can be coarsely inferred from the tracking output, thanks to the audio-feature-to-person assignment variables. The speaker diarization results obtained with our method are compared with two other methods \cite{vijayasenan2012diartk,gebru2018a} based on the Diarization Error Rate (DER) score.
}

The remainder of the paper is organized as follows. Section \ref{sec:related-work} describes the related work. Section \ref{sec:model} describes in detail the proposed  formulation. Section~\ref{sec:var-approx} describes the proposed variational approximation and Section~\ref{eq:vem} details the variational expectation-maximization procedure. The algorithm implementation is described in Section~\ref{sec:implementation}. Tracking results and comparisons with other methods are reported in Section~\ref{sec:experiments}. Finally, Section \ref{sec:conclusions} draws a few conclusions.\footnote{Supplemental materials are available at \url{https://team.inria.fr/perception/research/var-av-track/}}

\section{Related Work}
\label{sec:related-work}

In computer vision, there is a long history of multiple object tracking methods. While these methods provide interesting insights concerning the problem at hand, a detailed account of existing visual trackers is beyond the scope of this paper. 
Several audio-visual tracking methods were proposed in the recent past, e.g. 
\cite{checka04multiple,gatica2007audiovisual,hospedales08structure,NavqiMiaoChambers2010}. These papers proposed to use approximate inference of the filtering distribution using Markov chain Monte Carlo particle filter sampling (MCMC-PF).  These methods cannot provide estimates of the accuracy and merit of each modality with respect to each tracked person. 

More recently, audio-visual trackers based on particle filtering  and probability hypothesis density (PHD) filters were proposed, e.g. \cite{kilicc2015audio,schult2015information,barnard2016mean,kilicc2016mean,liu2017particle,liu2018non,qian20173d}. 
\cite{barnard2016mean} used DOAs of audio sources to guide the propagation of particles, and combined the filter with a mean-shift algorithm to reduce the computational complexity. Some PHD filter variants were proposed to improve the tracking performance \cite{liu2017particle,liu2018non}. The method of \cite{kilicc2015audio} also used DOAs of active audio sources to give more importance to particles located around DOAs. Along the same line of thought, \cite{kilicc2016mean} proposed a mean-shift sequential Monte Carlo PHD (SMC-PHD) algorithm that used audio information to improve the performance of a visual tracker. This implies that the persons being tracked must emit acoustic signals continuously and that multiple-source audio localization is reliable enough for proper audio-visual alignment.

PHD-based tracking methods are computationally efficient but their inherent limitation is that they are unable to associate observations to tracks. 
Hence they require an external post-processing mechanism that provides associations. Also, in the case of PF-based audio-visual filtering, the number of tracked persons must 
be set in advance and sampling can be a computational burden. 
In contrast, the proposed 
variational formulation embeds association variables within the model, uses a birth process to estimate the initial number of persons and to add new ones along 
time, and an explicit dynamic model yields smooth trajectories.

Another limitation of the methods proposed in \cite{gatica2007audiovisual,NavqiMiaoChambers2010,barnard2016mean,liu2017particle,liu2018non,qian20173d} 
is that they need as input a continuous flow of audio and visual observations. To some extent, this is also the case with \cite{kilicc2015audio,kilicc2016mean}, where only the audio observations are supposed to be continuous.
All these methods
showed good performance in the case of the AV16.3 dataset \cite{lathoud2004av16} in which the participants spoke simultaneously and continuously -- which is somehow artificial. The AV16.3 dataset was recorded in a specially equipped meeting room using three cameras that generally guarantee that frontal views of the participants were always available. This contrasts with the AVDIAR dataset which was recorded with one sensor unit composed of two cameras and six microphones. The AVDIAR scenarios are composed of participants that take speech turns while they look at each other, hence they speak intermittently and they do not always face the cameras. 

Recently, we proposed an audio-visual clustering method \cite{gebru2016em} and an audio-visual speaker diarization method \cite{gebru2018a}. The weighted-data clustering method of \cite{gebru2016em} analyzed a short time window composed of several audio and visual frames and hence it was assumed that the speakers were static within such temporal windows. Binaural audio features were mapped onto the image plane and were clustered with nearby visual features. There was no dynamic model that allowed to track speakers. The audio-visual diarization method \cite{gebru2018a} used an external multi-object visual tracker that provided trajectories for each tracked person. The audio-feature-space to image-plane mapping \cite{deleforge2015colocalization} was used to assign audio information to each tracked person at each time step. Diarization itself was modeled with a binary state variable (speaking or silent) associated with each person. The diarization transition probabilities (state dynamics) were hand
crafted, with the assumption that the speaking status of a person was independent of all the other persons. Because of the small number of state configurations, i.e. $\{0,1\}^N$ (where $N$ is the maximum number of tracked persons), the MAP solution could be found by exhaustively searching the state space. In Section~\ref{sec:diarization results} we use the AVDIAR recordings to compare our diarization results with the results obtained with \cite{gebru2018a}.

The variational Bayesian inference method proposed in this paper may well be viewed as a multimodal generalization of variational expectation maximization algorithms for multiple object tracking using either visual-only information \cite{ba2016line} or audio-only information \cite{li:hal-01851985,ban:hal-01969050}. We show that these models can be extended to deal with observations living in completely different mathematical spaces. Indeed, we show that two (or several) different data-processing pipelines can be embedded and treated on an equal footing in the proposed formulation. 
Special attention is given to audio-visual alignment and to audio-to-person assignments: (i)~we learn a mapping from the space of audio features to the image plane, as well as the inverse of this mapping, which are integrated in the proposed generative approach, and (ii) we show that the an increase in the number of assignment variables, due to the use of two modalities, do not affect the complexity of the algorithm. Absence of observed data of any kind or erroneous data are carefully modeled: this enables the algorithm to deal with intermittent observations, whether audio, visual, or both. This is probably one of the most prominent features of the method, in contrast with most existing audio-visual tracking methods which require continuous and simultaneous flows of visual and audio data.

This paper is an extended version of \cite{ban2017exploiting} and of \cite{ban2018accounting}. The probabilistic model and its variational approximation were briefly presented in \cite{ban2017exploiting} together with preliminary results obtained with three AVDIAR sequences. Reverberation-free audio features were used in \cite{ban2018accounting} where it was shown that good performance could be obtained with these features when the audio mapping was trained in one room and tested in another room. With respect to these two papers. we provide detailed descriptions of the proposed formulation, of the variational expectation maximization solver and of the implemented algorithm. We explain in detail the birth process, which is crucial for track initialization and for detecting potentially new tracks at each time step. We experiment with the entire AVDIAR dataset and we several sequences from the AV16.3 dataset; we benchmark our method with the state-of-the-art multiple-speaker audio-visual tracking methods \cite{kilicc2015audio,kilicc2016mean} and with \cite{ba2016line}. Moreover, we show that our tracker can be used for audio-visual speaker diarization \cite{gebru2018a}.

\section{Proposed Model}
\label{sec:model}
\subsection{Mathematical Definitions and Notations}
\label{subsec:defnot}
Unless otherwise specified, uppercase letters denote random variables while lowercase letters denote their realizations, \eg $p(X=x)$, where $p(\cdot)$ denotes either a probability density function (pdf) or a probability mass function (pmf). For the sake of conciseness we generally write $p(x)$. 
Vectors are written in slanted bold, \eg $\Xvect, \xvect$, whereas matrices are written in bold, \eg $\Ymat, \ymat$. 
Video and audio data are assumed to be synchronized, and let $t$ denote the common frame index. Let $N$ be the upper bound of the number of persons that can simultaneously be tracked at any time $t$, and let $n\in\{1\dots N\}$ be the person index. Let $n=0$ denote \textit{nobody}. 
A $t$ subscript denotes variable concatenation at time $t$, e.g. $\Xmat_t = (\Xvect_{t1}, \dots, \Xvect_{tn}, \dots, \Xvect_{tN})$, and  the subscript ${1:t}$ denotes concatenation from 1 to $t$, e.g. $\Xmat_{1:t} = (\Xmat_1, \dots, \Xmat_t)$.

Let $\Xvect_{tn}\in\mathcal{X}\subset\mathbb{R}^2$, $\Yvect_{tn}\in\mathcal{Y}\subset\mathbb{R}^2$ and $\Wvect_{tn}\in\mathcal{W}\subset\mathbb{R}^2$ be three latent variables that correspond to the 2D position, 2D velocity and 2D size (width and height) of person $n$ at $t$, respectively. Typically, $\Xvect_{tn}$ and $\Wvect_{tn}$ are the center and size of a bounding box of a person while $\Yvect_{tn}$ is the velocity of $\Xvect_{tn}$.  Let $\Svect_{t}=\{(\Xvect_{tn}^\top, \Wvect_{tn}^\top,\Yvect_{tn}^\top)^\top\}_{n=1}^{N}\subset\mathbb{R}^6$ be the complete set of continuous latent variables at $t$, where $\tp$ denotes the transpose operator. Without loss of generality, in this paper a person is characterized with the bounding box of her/his head and the center of this bounding box is assumed to be the location of the corresponding speech source.

We now define the observations. 
\addnote[observations]{1}{
At each time $t$ there are $M_t$ visual observations and $K_t$ audio observations. Let $\fmat_t = \{\fvect_{tm}\}_{m=1}^{M_t}$ and $\gmat_t = \{\gvect_{tk}\}_{k=1}^{K_t}$ be realizations of the visual and audio observed random variables $\{\Fvect_{tm}\}_{m=1}^{M_t}$ and $\{\Gvect_{tk}\}_{k=1}^{K_t}$, respectively}. Visual observations, $\fvect_{tm}= (\vvect_{tm}^\top, \uvect_{tm}^\top)^\top$, correspond to the bounding boxes of detected faces, namely the concatenation of the bounding-box center, width and height, $\vvect_{tm}\in \mathcal{V}\subset \mathbb{R}^4$, and of a feature vector $\uvect_{tm}\in \mathcal{H} \subset \mathbb{R}^{d}$ that describes the photometric content of that bounding box, i.e. a $d$-dimensional face descriptor (Section~\ref{subsec:visual_processing}). 
Audio observations, $\gvect_{tk}$, correspond to inter-channel spectral features, where $k$ is a frequency sub-band index. Let's assume that there are $K$ sub-bands, that  $K_t\leq K$ sub-bands are \textit{active} at $t$, i.e. sub-bands with sufficient signal energy, and that there are $J$ frequencies per sub-band. Hence, $\gvect_{tk}\in\mathbb{R}^{2J}$ corresponds to the real and imaginary parts of $J$ complex-valued Fourier coefficients. It is well established that inter-channel spectral features $\{\gvect_{tk}\}_{k=1}^{K_t}$ contain audio-source localization information, which is what is needed for tracking.
These audio features are obtained by applying the multi-channel audio processing method described in Section~\ref{subsec:audio_processing} below.
Note that both the number of visual and of audio observations at $t$, $M_t$ and $K_t$, vary over time. Let $\omat_{1:t}= (\omat_1, \dots, \omat_t)$ denote the set of observations from 1 to $t$, where $\omat_t = (\fmat_t, \gmat_t)$.

Finally, we define the assignment variables of the proposed latent variable model. There is an assignment variable (a discrete random variable) associated with each observed variable. Namely, let $A_{tm}$ and $B_{tk}$ be associated with $\fvect_{tm}$ and with $\gvect_{tk}$, respectively, e.g. $p(A_{tm}=n)$ denotes the probability of assigning visual observation $m$ at $t$ to person $n$. Note that $p(A_{tm}=0)$ and $p(B_{tk}=0)$ are the probabilities of assigning visual observation $m$ and audio observation $k$ to none of the persons, or to nobody. In the visual domain, this may correspond to a false detection while in the audio domain this may correspond to an audio signal that is not uttered by a person. 
There is an additional assignment variable, $C_{tk}$ that is associated with the audio generative model described in Section~\ref{subsec:audio-model}. 
The assignment variables are jointly denoted with $\Zvect_t=(\Avect_t, \Bvect_t, \Cvect_t)$.

\subsection{The Filtering Distribution}


We remind that the objective is to estimate the positions and velocities of participants (multiple person tracking) and, possibly, to estimate their speaking status (speaker diarization). 
The audio-visual multiple-person tracking problem is cast into the problems of estimating the filtering distribution $p(\smat_t, \zvect_t | \omat_{1:t})$ and of inferring the state variable $\Smat_t$. Subsequently, speaker diarization can be obtained from audio-feature-to-person information via the estimation of the assignment variables $\Bvect_{tk}$ (Section~\ref{sec:diarization}).

We reasonably assume that the state variable $\Smat_t$ follows a first-order Markov model, and that the visual and audio observations only depend on $\Smat_t$ and $\Zvect_t$. 
By applying Bayes rule, one can then write the filtering distribution of $(\smat_t, \zvect_t)$ as:
\begin{align}
p(\smat_t, \zvect_t | \omat_{1:t}) \propto p(\omat_t | \smat_t, \zvect_t) p(\zvect_t | \smat_t) p(\smat_t | \omat_{1:t-1}),
\label{eq:posterior_Bayes}
\end{align}
with:
\begin{align}
\label{eq:Bayes_observation}
p(\omat_t| \smat_t,\zvect_t) &= p(\fmat_t |  \smat_t,\avect_t) p(\gmat_t |  \smat_t,\bvect_t, \cvect_t),\\
\label{eq:Bayes_assign}
p(\zvect_t | \smat_t) &= p(\avect_t) p(\bvect_t) p(\cvect_t|\svect_t,\bvect_t),\\
\label{eq:predictive distribution}
p(\smat_t | \omat_{1:t-1}) &= \int p(\smat_t | \smat_{t-1}) p( \smat_{t-1} | \omat_{1:t-1}) d\smat_{t-1}.
\end{align}
Eq. \eqref{eq:Bayes_observation} is the joint (audio-visual) observed-data likelihood. Visual and audio observations are assumed independent conditionally to $\Smat_t$, and their distributions will be detailed in Sections~\ref{subsec:visual-model} and \ref{subsec:audio-model}, respectively.\footnote{We will see that $\Gmat_t$ depends on  $\Xmat_t$ but depends neither on $\Wmat_t$ nor on $\Ymat_t$, and $\Fmat_t$ depends on $\Xmat_t$ and $\Wmat_t$ but not on $\Ymat_t$.} Eq. \eqref{eq:Bayes_assign} is the prior distribution of the assignment variable. 
The observation-to-person assignments are assumed to be a priori independent so that the probabilities in \eqref{eq:Bayes_assign} factorize as: 
\begin{align}
\label{eq:priors-a}
p(\avect_t) &= \prod_{m=1}^{M_t} p(a_{tm}), \\
\label{eq:priors-b}
p(\bvect_t) & = \prod_{k=1}^{K_t} p(b_{tk}), \\
\label{eq:priors-c}
p(\cvect_t | \smat_t, \bvect_t) &= \prod_{k=1}^{K_t} p(c_{tk}|\svect_{tn}, B_{tk}=n). 
\end{align}
It makes sense to assume that these distributions do not depend on $t$ and that they are uniform. 
The following notations are introduced: $\eta_{mn} = p(A_{tm}=n)=1/(N+1)$ and $\rho_{kn} = p(B_{tk}=n)=1/(N+1)$. 
The probability $p(c_{tk}|\svect_{tn}, B_{tk}=n)$ is discussed below (Section~\ref{subsec:audio-model}).

Eq. \eqref{eq:predictive distribution} is the predictive distribution of $\smat_t$ given the past observations, i.e. from 1 to $t-1$. The state dynamics 
in \eqref{eq:predictive distribution} are modeled with a linear-Gaussian first-order Markov process. Moreover, it is assumed that the dynamics are independent 
over speakers:
\begin{equation}
\label{eq:state_dynamic}
p(\smat_t | \smat_{t-1}) = \prod_{n=1}^{N} \mathcal{N}
(\svect_{tn};\Dmat \svect_{t-1\: n},\Lambdamat_{tn}),
\end{equation}
where $\Lambdamat_{tn}$ is the dynamics’ covariance matrix and $\Dmat$ is the state transition matrix, given by:
\begin{center}
$\Dmat = 
\begin{pmatrix}
 \multirow{2}{*}{$\Imat_{4\times4}$} & \Imat_{2\times2}\\
 & \textbf{0}_{2\times2}\\
 \textbf{0}_{2\times4} & \Imat_{2\times2}\\
\end{pmatrix}.
$
\end{center}
As described in Section~\ref{sec:var-approx} below, an important feature of the proposed model is that the predictive distribution \eqref{eq:predictive distribution} at frame $t$ is computed from the state dynamics model \eqref{eq:state_dynamic} and an approximation of the filtering distribution $p(\smat_{t-1} | \omat_{1:t-1})$ at frame $t-1$, which also factorizes across speaker. As a result, the computation of \eqref{eq:predictive distribution} factorizes across speakers as well.

\subsection{The Visual Observation Model}
\label{subsec:visual-model}

As already mentioned above (Section~\ref{subsec:defnot}), a visual observation $\fvect_{tm}$ consists of the center, width and height of a bounding box, namely $\vvect_{tm}\in \mathcal{V}\subset \mathbb{R}^4$, as well as of a feature vector $\uvect_{tm}\in \mathcal{H} \subset \mathbb{R}^{d}$ describing the region inside the bounding box. Since the velocity is not observed, 
a $4\times6$ projection matrix $\Pmat_{f}=(\Imat_{4\times4} \; \textbf{0}_{4\times2})$ is used to project $\svect_{tn}$ onto $\mathcal{V}$.
\addnote[appearance]{2}{Assuming that the $M_t$ visual observations $\{ \fvect_{tm} \}_{m=1}^{M_t}$ available at $t$ are independent, and that the appearance representation of a person is independent of his/her position in the image, e.g. CNN-based embedding, the visual likelihood in \eqref{eq:Bayes_observation} is defined as:}
\begin{align}
\label{eq:visual_model}
p(\fmat_t |  \smat_t, \avect_t) =  \prod_{m=1}^{M_t} p(\vvect_{tm}|\smat_t, a_{tm}) p(\uvect_{tm}|\hmat,a_{tm}),
\end{align}
where the observed bounding-box centers, widths, heights, and feature vectors are drawn from the following distributions:
\begin{align}
\label{eq:bb-model}
p(\vvect_{tm} & |\smat_t,A_{tm}=n)  = 
\begin{cases}
\mathcal{N} (\vvect_{tm}; \Pmat_{f}\svect_{tn}, \Phimat_{tm}) & \textrm{if} \: 1\leq n \leq N \\
 \mathcal{U} ( \vvect_{tm} ; \textrm{vol}(\mathcal{V})) & \textrm{if} \: n=0,
 \end{cases} 
 \\
\label{eq:app-model}
p(\uvect_{tm} & |\hmat,A_{tm}=n) =
\begin{cases}
\mathcal{B} (\uvect_{tm}; \hvect_{n}) & \textrm{if} \: 1\leq n \leq N \\
 \mathcal{U} ( \uvect_{tm}; \textrm{vol}(\mathcal{H})) & \textrm{if} \: n=0,
 \end{cases} 
\end{align}
where $\Phimat_{tm}\in\mathbb{R}^{4 \times 4}$ is a covariance matrix quantifying the measurement error in the bounding-box center and size, $\mathcal{U}(\cdot 
; \textrm{vol}(\cdot))$ is the uniform distribution with $\textrm{vol}(\cdot)$ being the volume of the support of the variable, $\mathcal{B} (\cdot; 
\hvect)$ is the \addnote[bhattacharya]{1}{Bhattacharya distribution \cite{bhattacharyya1943measure}}, 
and $\hmat = (\hvect_1, \dots, \hvect_N) \in \mathbb{R}^{d \times N}$ is a set of prototype feature vectors that model the appearances of the $N$ persons.

\subsection{The Audio Observation Model}
\label{subsec:audio-model}

It is well established in the recent audio signal processing literature that inter-channel spectral features encode sound-source localization information \cite{deleforge2015colocalization,dorfan2015tree,li2016estimation}. Therefore, observed audio features, $\gmat_t=\{\gvect_{tk}\}_{k=1}^{K_t}$ are obtained by considering all the pairs of a microphone array.
Audio observations depend neither on the size of the bounding box $\wvect_t$, nor on the velocity $\yvect_t$. 
\addnote[audio-observations-1]{2}{Indeed, we note that the velocity of a sound source (a moving person) is of about 1 meter/second, which is negligible compared to the speed of sound. Moreover, the inter-microphone distance is small compared to the source-to-microphone distance, hence the Doppler effect, if any, is similar across microphones}. Hence one can replace $\svect$ with $\xvect=\Pmat_g \svect$ in the equations below, with $\Pmat_g=(\Imat_{2\times 2} \; \zeromat_{2\times 4})$.
By assuming independence across frequency sub-bands (indexed by $k$), the audio likelihood in \eqref{eq:Bayes_observation} can be factorized as:
\begin{equation}
\label{eq:audio-independence}
p(\gmat_t |  \smat_t,\bvect_t, \cvect_t) = 
\prod_{k=1}^{K_t} p (\gvect_{tk} | \xvect_{tb_{tk}}, b_{tk}, c_{tk}).
\end{equation}
While the inter-channel spectral features $\gvect_{tk}$ contain localization information, in complex acoustic environments there is no explicit transformation 
that maps a source location onto an inter-channel spectral feature. We therefore make recourse to modeling this mapping via learning a non-linear regression. 
We use the method of \cite{li2016estimation} to extract audio features and the piecewise-linear regression model of \cite{DeleforgeForbesHoraud2015} to learn a 
mapping between the space of audio-source locations and the space of audio features. The method of \cite{DeleforgeForbesHoraud2015}
belongs to the mixture of experts (MOE) class of models and hence it embeds well in our latent-variable mixture model. Let $\{h_{kr}\}_{r=1}^{r=R}$ be a set of  linear regressions, such that the $r$-th linear transformation $h_{kr}$ maps $\xvect\in\mathbb{R}^2$ onto $\gvect_k\in\mathbb{R}^{2J}$ for the frequency sub-band $k$. It follows that \eqref{eq:audio-independence} writes:
\begin{align}
\label{eq:affine-mapping}
p (  \gvect_{tk} |&  \xvect_{tn},  B_{tk} =n , C_{tk} = r)  =  \\
&
\begin{cases}
\mathcal{N} (\gvect_{tk} ; h_{kr} (\xvect_{tn}), \Sigmamat_{kr}) & \textrm{if} \: 1\leq n \leq N \\
\mathcal{U} (\gvect_{tk} ; \textrm{vol} (\mathcal{G})) & \textrm{if} \: n=0,
\end{cases} \nonumber
\end{align}
where $\Sigmamat_{kr}\in \mathbb{R}^{2J\times 2J}$ is a covariance matrix that captures the linear-mapping error and $C_{tk}$ is a discrete random variable, 
such that $C_{tk} = r$ means that the audio feature $\gvect_{tk}$ is generated through the $r$-th linear transformation.
\addnote[appendix-ref]{1}{
Please consult Appendix~\ref{app:gllim} for details on how the parameters of the linear transformations $h_{kr}$ are learned from a training dataset.
}

\section{Variational Approximation}
\label{sec:inference}
\label{sec:var-approx}

Direct estimation of the filtering distribution $p(\smat_t, \zvect_t | \omat_{1:t})$ is computationally intractable. %
Consequently, evaluating expectations over this 
distribution is intractable as well. We overcome this problem via variational inference and associated EM closed-form solver \cite{bishop06pa,smidl06var}. More 
precisely $p(\smat_t, \zvect_t | \omat_{1:t})$ is approximated with the following factorized form:
\begin{equation}
\label{eq:variational-approximation}
p(\smat_t,\zvect_t | \omat_{1:t}) \approx q(\smat_t,\zvect_t) = q(\smat_t)q(\zvect_{t}), 
\end{equation} 
which implies
\begin{equation}
q(\smat_t) = \prod_{n=1}^{N} q(\svect_{tn}), \; q(\zvect_t) = \prod_{m=1}^{M_t} q (a_{tm}) \prod_{k=1}^{K} q(b_{tk},c_{tk}),
\end{equation}
where $q(A_{tm} = n)$ and $q(B_{tk} = n, C_{tk} =r)$ are the variational posterior probabilities of assigning visual observation $m$ to person $n$ and audio observation $k$ to person $n$, respectively. The proposed variational approximation \eqref{eq:variational-approximation} amounts to break the conditional dependence of $\Smat$ and $\Zvect$ with respect to $\omat_{1:t}$ which causes the computational intractability. 
Note that the visual, $\Avect_{t}$, and audio, $\Bvect_{t}$, $\Cvect_{t}$,  assignment variables are independent, that the assignment variables for each observation are also independent, and that $B_{tk}$ and $C_{tk}$ are conditionally dependent on the audio observation.
This factorized approximation makes the calculation of $p(\smat_t, \zvect_t | \omat_{1:t})$ tractable. The optimal solution 
is given by an instance of the variational expectation maximization (VEM) algorithm \cite{bishop06pa,smidl06var}, which alternates between two steps: 
\begin{itemize}
\item \textit{Variational} E-step: the approximate log-posterior distribution of each one of the latent variables is estimated by taking the expectation of the complete-data log-likelihood over the remaining latent variables, i.e. \eqref{eq:variational-state}, \eqref{eq:variational-assignment-1}, and \eqref{eq:variational-assignment-2} below, and
\item M-step: model parameters are estimated by maximizing the variational expected complete-data log-likelihood.\footnote{Even if the M-step is in closed-form, 
the inference is based on the variational posterior distributions. Therefore, the M-step could also be regarded as \textit{variational}.}
\end{itemize}
In the case of the proposed model the latent variable log-posteriors write:
\begin{align}
\label{eq:variational-state}
 \log  q  & (\svect_{tn})  = \mathbb{E}_{q(\zvect_{t}) \prod_{\ell \neq n} 
q(\svect_{t\ell})} [\log p(\smat_{t}, \zvect_{t} | \omat_{1:t} )] + \mbox{const},\\
\label{eq:variational-assignment-1}
 \log  q   &  (a_{tm})   =   \\
 & \mathbb{E}_{q(\mathbf{s}_t)\prod_{\ell \neq m} q(a_{t\ell})\prod_{k} q(b_{tk},c_{tk})}  [\log p(\smat_{t}, \zvect_{t} | \omat_{1:t})] + \mbox{const}, \nonumber \\
\label{eq:variational-assignment-2}
 \log  q &  (b_{tk},c_{tk})  =  \\
& \mathbb{E}_{q(\mathbf{s}_t)\prod_{m} q(a_{tm})\prod_{\ell \neq k} q(b_{t\ell},c_{t\ell})}  [\log p(\smat_{t}, \zvect_{t} | \omat_{1:t})] + \mbox{const}. \nonumber
\end{align}
A remarkable consequence of the factorization~\eqref{eq:variational-approximation} is that $p(\smat_{t-1}|\omat_{1:t-1})$ is replaced with $q(\smat_{t-1})=\prod_{n=1}^{N} q(\svect_{t-1 \: n})$, consequently \eqref{eq:predictive distribution} becomes:
\begin{align}
\label{eqn:predictive-motion}
 p(\smat_{t}|\omat_{1:t-1}) & \approx \int    p(\smat_{t}|\smat_{t-1})\prod_{n=1}^N q(\svect_{t-1\: n})d\smat_{t-1}. 
\end{align}
It is now assumed that the variational posterior distribution $q(\svect_{t-1\ n})$ is Gaussian with mean $\muvect_{t-1\ n}$ and covariance $\Gammamat_{t-1\ n}$: 
\begin{equation}
\label{eq:var-posterior}
q(\svect_{t-1\ n}) = \mathcal{N}(\svect_{t-1\ n}; \muvect_{t-1\ n},\Gammamat_{t-1\ n}).
\end{equation}
By substituting \eqref{eq:var-posterior} into~\eqref{eqn:predictive-motion} and combining it with \eqref{eq:state_dynamic}, the predictive distribution \eqref{eqn:predictive-motion} becomes:
\begin{align}
\label{eqn:predictive-motion-mod}
p(\smat_{t}|\omat_{1:t-1}) \approx  \prod_{n=1}^{N}\mathcal{N}(\svect_{tn};\Dmat\muvect_{t-1\: n},\Dmat\Gammamat_{t-1\: n}\Dmat^{\top}+\Lambdamat_{tn}).
\end{align}
Note that the above distribution factorizes across persons. Now that all the factors in \eqref{eq:posterior_Bayes} have tractable expressions, a VEM algorithm 
can be derived.

\section{Variational Expectation Maximization}
\label{eq:vem}
The proposed VEM algorithm iterates between an E-S-step, an E-Z-step, and an M-step on the following grounds.

\subsubsection{E-S-step}  the per-person variational posterior distribution of the state vector $q(\svect_{tn})$ is evaluated by developing 
\eqref{eq:variational-state}. The joint posterior $p(\smat_{t}, \zvect_{t} | \omat_{1:t} )$ in \eqref{eq:variational-state} is the product of 
\eqref{eq:Bayes_observation}, \eqref{eq:Bayes_assign} and \eqref{eqn:predictive-motion-mod}. We thus first sum the logarithms of \eqref{eq:Bayes_observation}, 
of \eqref{eq:Bayes_assign} and of \eqref{eqn:predictive-motion-mod}. Then we ignore the terms that do not involve $\svect_{tn}$. Evaluation of the expectation 
over all the latent variables except $\svect_{tn}$ yields the following Gaussian distribution: 
\begin{equation}
\label{eq:var-posterior-t}
q(\svect_{t n}) = \mathcal{N}(\svect_{t n}; \muvect_{tn},\Gammamat_{t n}),
\end{equation}
with:
\begin{align}
\label{eq:posteriordistribution-of-Xtn-cov}
& \Gammamat_{tn}   =  \Bigg(  
\underbrace{\sum_{k=1}^{K}\sum_{r=1}^{R}  \beta_{tknr} \Pmat_{g}^{\top}\Lmat_{kr}^{\top} {\Sigmamat_{kr}\inv} \Lmat_{kr} \Pmat_{g} }_{\#1} \\
& + \underbrace{\sum_{m=1}^{M_t}\alpha_{tmn} {\Pmat_{f}^{\top} \Phimat_{tm}\inv\Pmat_{f}}}_{\#2}    
+ \underbrace{ \Big(\Lambdamat_{tn}+\Dmat\Gammamat_{t-1 \: n}\Dmat^{\top}\Big)\inv}_{\#3} \Bigg)\inv,  \nonumber
\end{align}
and with:
\begin{align}
\label{eq:posterior-distribution-of-Xtn-mean}
& \muvect_{tn}    =   \Gammamat_{tn}  \Bigg(  \underbrace{\sum_{k=1}^{K}\sum_{r=1}^{R} \beta_{tknr} \Pmat_{g}^{\top} \Lmat_{kr}^{\top} {\Sigmamat_{kr}\inv}(\gvect_{kr}-{\lvect_{kr}})}_{\#1}
  \\ 
& + \underbrace{ \sum_{m=1}^{M_t}\alpha_{tmn} {\Pmat_{f}^{\top} \Phimat_{tm}\inv} \vvect_{tm} }_{\#2}
 + \underbrace{\Big(\Lambdamat_{tn}+\Dmat\Gammamat_{t-1 \: n}\Dmat^{\top}\Big)\inv\Dmat\muvect_{t-1 \: n}  }_{\#3}
\Bigg), \nonumber
\end{align}
where $\alpha_{tmn} = q(A_{tm} = n)$ and $\beta_{tknr} = q(B_{tk} = n, C_{tk} =r)$ are computed in the E-Z-step below. A key point is that, because of the 
recursive nature of the formulas above, it is sufficient to make the Gaussian assumption at $t=1$, i.e. $q(\svect_{1n}) = 
\mathcal{N}(\svect_{1n};\muvect_{1n},\Gammamat_{1n})$, whose parameters may be easily initialized. It follows that $q(\svect_{tn})$ is Gaussian at every frame.

We note that both \eqref{eq:posteriordistribution-of-Xtn-cov} and \eqref{eq:posterior-distribution-of-Xtn-mean} are composed of three terms: the first (\#1), second (\#2) and third terms (\#3) of \eqref{eq:posteriordistribution-of-Xtn-cov} 
and of \eqref{eq:posterior-distribution-of-Xtn-mean} correspond to the audio, visual, and past cumulated information contributions to the precision matrix and 
the mean vector, respectively. Remind that the covariance $\Phimat_{tm}$ is associated with the visual observed variable in \eqref{eq:bb-model}. 
Matrices $\Lmat_{kr}$ and vectors $\lvect_{kr}$ characterize the piecewise affine mappings from the space of person locations to the space of audio features, i.e. Appendix~\ref{app:gllim}, and covariances $\Sigmamat_{kr}$ capture the errors that are associated with both audio measurements and the piecewise affine approximation in \eqref{eq:affine-mapping}. 
A similar interpretation holds for the three terms of \eqref{eq:posterior-distribution-of-Xtn-mean}.
\subsubsection{E-Z-step}
by developing \eqref{eq:variational-assignment-1}, along the same reasoning as above, we obtain the following closed-form expression for the variational posterior distribution of the visual assignment variable:
\begin{equation}
\label{eq:posterior distribution of visual assignment}
\alpha_{tmn}= q(A_{tm} = n) =  \frac{\tau_{tmn} \eta_{mn}}
{\sum_{i=0}^{N}\tau_{tmi} \eta_{mi}},
\end{equation}
where $\tau_{tmn}$ is given by:
\begin{align}
\tau_{tmn}  =  
\begin{cases}
\mathcal{N}(\vvect_{tm}; \Pmat_{f} \muvect_{tn}, \Phimat_{tm}) 
e^{-\frac{1}{2} \textrm{tr} 
\left( 
\Pmat_{f}^{\top} \Phimat_{tm}^{-1} \Pmat_{f}
\Gammamat_{tn}\right)} \\
\times \  \mathcal{B} (\uvect_{tm};\hvect_{n})   \quad \qquad \qquad \qquad \qquad \textrm{if} \: 1\leq n \leq N \\
 \mathcal{U}(\vvect_{tm}; \textrm{vol}(\mathcal{V})) \mathcal{U}(\uvect_{tm}; \textrm{vol}(\mathcal{H}))  \qquad \textrm{if} \: n =0.
\end{cases} \nonumber
\end{align}
Similarly, for the variational posterior distribution of the audio assignment variables, developing \eqref{eq:variational-assignment-2} leads to:
\begin{equation}
\beta_{tknr} = q(B_{tk} = n, C_{tk} =r) =  \frac{\kappa_{tknr} \rho_{kn} \pi_{r}}
{\sum_{i=0}^{N} \sum_{j=1}^{R} 
\kappa_{tkij} \rho_{ki} \pi_{j}
},\label{eq:posterior-audio}
\end{equation}
where $\kappa_{tknr}$ is given by:
\begin{align}
\label{eq:posterior distribution of audio assignment}
& \kappa_{tknr}  = \\
&
\begin{cases}
 \mathcal{N}(\gvect_{tk};\Lmat_{kr}\Pmat_{g}\muvect_{tn} + {\lvect_{kr}},\Sigmamat_{kr})
 e^{-\frac{1}{2}\textrm{tr}\left(\Pmat_{g}^{\top} \Lmat_{kr}^{\top} \Sigmamat_{kr}^{-1} \Lmat_{kr}\Pmat_{g} \Gammamat_{t n}\right)}  \\
\times \ \mathcal{N}(\tilde{\xvect}_{tn};\nuvect_{r}, \Omegamat_{r}))  \qquad \qquad  \textrm{if} \: 1\leq n \leq N \\
 \mathcal{U} (\gvect_{tk}; \textrm{vol}(\mathcal{G})) \qquad \qquad \qquad  \textrm{if} \: n =0.
\end{cases}
\nonumber
\end{align}
To obtain \eqref{eq:posterior distribution of audio assignment}, an additional approximation is made. Indeed, the logarithm of (39)
in Appendix~\ref{app:gllim}  is part of the complete-data log-likelihood and the denominator of this formula contains a weighted sum of Gaussian distributions. Taking the expectation of this term is not tractable because of the denominator. Based on the dynamical model \eqref{eq:state_dynamic}, we replace the state variable $\xvect_{tn}$ in (39) 
with a ``naive'' estimate $\tilde{\xvect}_{tn}$ predicted from the position and velocity inferred at $t-1$: $\tilde{\xvect}_{tn} = \xvect_{t-1\: n} + \yvect_{t-1\: n}$. 

\subsubsection{M-step} The entries of the covariance matrix of the state dynamics, $\Lambdamat_{tn}$, are the only parameters that need be estimated.
To this aim, we develop $ \mathbb{E}_{q(\mathbf{s}_{t})q(\zvect_{t})} [\log
 p(\smat_{t},\zvect_{t} | \omat_{1:t} )]$ and ignore the terms that do not depend on $\Lambdamat_{tn}$. We obtain: 
\begin{align*}
J(\Lambdamat_{tn}) = 
\mathbb{E}_{q(\svect_{tn})}\big[
\log \mathcal{N}(\svect_{tn}; \Dmat \muvect_{t-1\: n}, \Dmat \Gammamat_{t-1\: n} \Dmat^{\top} + \Lambdamat_{tn}) 
\big],
\end{align*}
which can be further developed as:
\begin{align}
\label{eq:mstep-lambda}
J & (\Lambdamat_{tn}) = 
\log | \Dmat \Gammamat_{t-1 \: n} \Dmat^{\top} + \Lambdamat_{tn} | + 
\textrm{Tr} \big((\Dmat \Gammamat_{t-1\: n} \Dmat^{\top} + \Lambdamat_{tn})^{-1} \nonumber \\
&\times \left( (\muvect_{tn} - \Dmat \muvect_{t-1\: n})
(\muvect_{tn} - \Dmat \muvect_{t-1\: n})^{\top} +
\Gammamat_{tn} \right) \big).
\end{align}
Hence, by differentiating \eqref{eq:mstep-lambda} with respect to $\Lambdamat_{tn}$ and equating to zero, we obtain:
\begin{equation}
\label{eq:M-steps}
\Lambdamat_{tn} = \Gammamat_{tn} - \Dmat \Gammamat_{t-1\: n} \Dmat^{\top} + (\muvect_{tn} - \Dmat \muvect_{t-1\: n})
(\muvect_{tn} - \Dmat \muvect_{t-1\: n})^{\top}.
\end{equation}

\section{Algorithm Implementation}
\label{sec:implementation}

The VEM procedure above will be referred to as VAVIT which stands for \textit{variational audio-visual tracking}, and pseudo-code is shown in 
Algorithm~\ref{algo:av-tracking}. In theory, the order in which the two expectation steps are executed is not important. In practice, the issue of 
initialization is crucial. In our case, it is more convenient to start with the E-Z step rather than with the E-S step because the former is easier to 
initialize than the latter (see below). We start by explaining how the algorithm is initialized at $t=1$ and then how the E-Z-step is initialized at each 
iteration. Next, we explain in detail the birth process. An interesting feature of the proposed method is that it allows to estimate who speaks when (i.e.\ 
perform speaker diarization) which is explained in detail at the end of the section.

\begin{figure}[t!]
\removelatexerror
\begin{algorithm}[H]
\caption{ \label{algo:av-tracking} Variational audio-visual tracking (VAVIT).}
 \KwIn{visual observations $\fmat_{1:t} = \{\vmat_{1:t},\ximat_{1:t}\}$\;
 \ \ \ \ \ \ \ \ \ audio observations $\gmat_{1:t}$\;
 }
 
 \KwOut{Parameters of $q(\smat_{1:t})$: $\{\muvect_{1:t,n},\Gammamat_{1:t,n}\}_{n=0}^N$ (the estimated position of each person $n$ is given by the two first entries of $\muvect_{1:t,n}$)\;
 \ \ \ \ \ \ \ \ \ \ \ Person speaking status for $1:t$}  
 Initialization (see Section \ref{sec:initialization})\;
 \For {$t = 1$ to end}
 {
    Gather visual and audio observations at frame $t$\;
    Perform voice activity detection\;
    Initialization of E-Z step (see Section \ref{sec:initialization});\\
	\For {$iter = 1$ to $N_{\text{iter}}$}
 	{
	E-Z-step (vision):\\ 
 	\For {$m \in \{1,...,M_t\}$}
 	{ \For {$n \in \{0,...,N_t\}$}{
   Evaluate $q(A_{tm} = n)$ with \eqref{eq:posterior distribution of visual assignment}\;
   }
   }
   E-Z-step (audio):\\ 
   \For {$k \in \{1,...,K_t\}$}
   {
   \For {$n \in \{0,...,N_t\}$ and $r \in \{1,...,R\}$ }
   {
   Evaluate $q(B_{tk} = n, C_{tk} =r)$ with \eqref{eq:posterior-audio} and \eqref{eq:posterior distribution of audio assignment}\;
   }
   }
   E-S-step:\\ 
   \For {$n \in \{1,...,N_t\}$ }
   {
   Evaluate $\Gammamat_{tn}$ and $\muvect_{tn} $ with \eqref{eq:posteriordistribution-of-Xtn-cov} and \eqref{eq:posterior-distribution-of-Xtn-mean};

   }

   M-step: Evaluate $\Lambdamat_{tn}$ with \eqref{eq:M-steps};
   }
   Perform birth (see Section \ref{sec:birth and death})\;
   Output the results\;
   }
\end{algorithm}
\end{figure}
\subsection{Initialization}
 \label{sec:initialization}


At $t=1$ one must provide initial values for the parameters of the distributions \eqref{eq:var-posterior-t}, namely $\muvect_{1n}$ and $\Gammamat_{1n}$ for all $n\in\{1\dots N\}$. These parameters are initialized as follows. The means are initialized at the image center and the covariances are given very large values, such that the variational distributions $q(\svect_{1n})$ are non-informative. Once these parameters are initialized, they remain constant for a few frames, i.e. until the birth process is activated (see Section~\ref{sec:birth and death} below).

As already mentioned, it is preferable to start with the E-Z-step than with the E-S-step because the initialization of the former is straightforward. Indeed, the E-S-step (Section~\ref{eq:vem}) requires current values for the posterior probabilities \eqref{eq:posterior distribution of visual assignment} and \eqref{eq:posterior distribution of audio assignment} which are estimated during the E-Z-step and which are both difficult to initialize. Conversely, the E-Z-step only requires current mean values, $\muvect_{tn}$, which can be easily initialized by using the model dynamics \eqref{eq:state_dynamic}, namely $\muvect_{tn}=\Dmat \muvect_{t-1 n}$.


\subsection{Birth Process}
\label{sec:birth and death}
We now explain in detail the birth process, which is executed at the start of the tracking to initialize a latent variable for each detected person, as well as at any time $t$ to detect new persons.
The birth process considers $B$ consecutive visual frames. 
At $t$, with $t>B$, we consider the set of visual observations assigned to $n=0$ from $t-B$ to $t$, namely observations whose posteriors \eqref{eq:posterior 
distribution of visual assignment} are maximized for $n=0$ (at initialization all the observations are in this case). We then build observation sequences from 
this set, namely sequences of the form $(\tilde{\vvect}_{m_{t-B}}, \dots, \tilde{\vvect}_{m_{t}})_{\tilde{n}} \in\mathcal{B}$, where $m_t$ indexes the set of 
observations at $t$ assigned to $n=0$ and $\tilde{n}$ indexes the set $\mathcal{B}$ of all such sequences. Notice that the birth process only uses the 
bounding-box center, width and size, $\vvect$, and that the descriptor $\uvect$ is not used. Hence the birth process is only based on the smoothness of an 
observed sequence of bounding boxes.
Let's consider the marginal likelihood of a sequence $\tilde{n}$, namely:
\begin{align}
\label{eqn:birth-test}
\mathcal{L}_{\tilde{n}} & =  p  ( (\tilde{\vvect}_{m_{t-B}},  \dots, \tilde{\vvect}_{m_{t}})_{\tilde{n}} )\\
= & \int \dots \int p(\tilde{\vvect}_{m_{t-B}} | \svect_{t-B\:\tilde{n}}) \dots p(\tilde{\vvect}_{m_{t}} | \svect_{t\: \tilde{n}}) \nonumber \\
 \times & p(\svect_{t\: \tilde{n}} | \svect_{t-1\:\tilde{n}}) \dots p(\svect_{t-B+1\: \tilde{n}} | \svect_{t-B\: \tilde{n}}) 
p(\svect_{t-B\:\tilde{n}}) d \svect_{t-B:t\:\tilde{n}}, \nonumber
\end{align}
where $\svect_{t,\tilde{n}}$ is the latent variable already defined and $\tilde{n}$ indexes the set $\mathcal{B}$. All the probability distributions in \eqref{eqn:birth-test} were already defined, namely \eqref{eq:state_dynamic} and \eqref{eq:bb-model}, with the exception of $p(\svect_{t-B,\tilde{n}})$. Without loss of generality, we can assume that the latter is a normal distribution centered at $\tilde{\vvect}_{m_{t}}$ and with a large covariance. Therefore, the evaluation of \eqref{eqn:birth-test} yields a closed-form expression for $\mathcal{L}_{\tilde{n}}$. A sequence $\tilde{n}$ generated by a person is likely to be smooth and hence $\mathcal{L}_{\tilde{n}}$ is high, while for a non-smooth sequence the marginal likelihood is low. A newborn person is therefore created from a sequence of observations $\tilde{n}$ if $\mathcal{L}_{\tilde{n}}>\tau$, where $\tau$ is a user-defined parameter. As just mentioned, the birth process is executed to initialize persons as well as along time to add new persons. In practice, in \eqref{eqn:birth-test} we set $B=3$ and hence, from $t=1$ to $t=4$ all the observations are initially assigned to $n=0$. 

\subsection{Speaker Diarization}
\label{sec:diarization}

Speaker diarization consists of assigning temporal segment of speech to persons \cite{anguera2012speaker}.
We introduce a binary variable $\chi_{tn}$ such that $\chi_{tn}=1$ if person $n$ speaks at time $t$ and $\chi_{tn}=0$ otherwise.
Traditionally, speaker diarization is based on the following assumptions. First, it is assumed that speech signals are sparse in the time-frequency domain. Second, it is assumed that each time-frequency point in such a spectrogram corresponds to a single speech source. Therefore, the proposed speaker diarization  method is based on assigning time-frequency points to persons. 

In the case of the proposed model, speaker diarization can be coarsely inferred from frequency sub-bands in the following way. The posterior probability that the speech signal available in the frequency sub-band $k$ at frame $t$ was uttered by person $n$, given the audio observation $\gvect_{tk}$, is:
\begin{align}
\label{eq:posterior diarization}
 p (B_{tk} = n | \gvect_{tk}) 
 &= \sum_{r = 1}^{R} p (B_{tk} = n, C_{tk} = r | \gvect_{tk}),
\end{align}
where $B_{tk}$ is the audio assignment variable and $C_{tk}$ is the affine-mapping assignment variable defined in Section~\ref{subsec:audio-model} and in Appendix~\ref{app:gllim}.
Using the variational approximation \eqref{eq:posterior-audio}, this probability becomes:
\begin{align}
\label{eq:variational posterior diarization}
 p (B_{tk} = n | \gvect_{tk})  &\approx \sum_{r = 1}^{R} q (B_{tk} = n, C_{tk} = r) \nonumber \\
 & =  \sum_{r = 1}^{R} \beta_{tknr},
\end{align}
and by accumulating probabilities over all the frequency sub-bands, we obtain the following formula:
\begin{equation}
\label{eq:diarization-decision}
\chi_{tn} =  
\begin{cases}
1 & \textrm{if} \quad \frac{1}{K_t}\sum_{k = 1}^{K_t} \sum_{r = 1}^{R} \beta_{tknr}   \geq \gamma \\
0 & \textrm{otherwise},
\end{cases}
\end{equation} 
where $\gamma$ is a user-defined threshold. Note that there is no dynamic model associated with diarization: $\chi_{tn}$ is estimated independently at each frame and for each person. More sophisticated diarization models can be found in \cite{noulas2012multimodal,gebru2018a}.

\section{Experiments}
\label{sec:experiments}

\subsection{The AVDIAR Dataset}

We used the AVDIAR\footnote{\url{https://team.inria.fr/perception/avdiar/}} dataset \cite{gebru2018a} 
to evaluate the performance of the proposed audio-visual tracking method. This dataset is challenging in terms of audio-visual analysis. There are several 
participants involved in informal conversations while wandering around. They are in between two and four meters away from the audio-visual recording device. 
They take speech turns and often there are speech overlaps. They turn their faces away from the camera. The dataset is annotated as follows: The visual 
annotations comprise the centers, widths and heights of two bounding boxes for each person and in each video frame, a face bounding box and an upper-body 
bounding box. An identity (a number) is associated with each person through the entire dataset. The audio annotations comprise the speech status of each person 
over time (speaking or silent), with a minimum speech duration of $0.2$~s. The audio source locations correspond to the centers of the face bounding 
boxes.

The dataset was recorded with a sensor composed of two cameras and six microphones, but only one camera is used in the experiments described below. The videos 
were recorded at $25$~FPS. The frame resolution is of $1920\times 1200$ pixels corresponding to a field of view of $97^{\circ} \times 80^{\circ}$. 
The microphone signals are sampled at $16$~kHz. The dataset was recorded into two different rooms, \textit{living-room} and \textit{meeting-room}, e.g. 
Fig.~\ref{fig:Seq13-4P-S2M1} and Fig.~\ref{fig:Seq19-2P-S1M1}. These two rooms have quite different lighting conditions and acoustic properties (size, presence 
of furniture, background noise, etc.). Altogether there are 18 sequences associated with living-room (26927 video frames) and 6 sequences with meeting-room 
(6031 video frames). Additionally, there are two training datasets, $\mathcal{T}_1$ and $\mathcal{T}_2$ (one for each room) that contain input-output pairs of 
multichannel audio features and audio-source locations that allow to estimate the parameters \eqref{eq:parameters-theta} using the method of 
\cite{deleforge2015colocalization}. This yields a mapping between source locations in the image plane, $\xvect$, and audio features, $\gvect$. Audio feature extraction is described in detail below.

One interesting characteristic of the proposed tracking is its flexibility in dealing only with visual data, only with audio data, or with visual and audio data. Moreover, the algorithm is able to automatically switch from unimodal (audio or visual) to multimodal (audio and visual). In order to quantitatively assess the performance and merits of each one of these variants we used two configurations:
\begin{itemize}
\item \textit{Full camera field of view (FFOV)}: The entire horizontal field of view of the camera, i.e. 1920 pixels, or 97$^\circ$, is being used, such that visual and audio observations, \textbf{if any}, are simultaneously available, and
\item \textit{Partial camera field of view (PFOV)}: The horizontal field of view is restricted to 768 pixels (or 49$^\circ$) and there are two \textit{blind} strips (576 pixels each) on its left- and right-hand sides; the \textit{audio field of view} remains unchanged, 1920 pixels, or 97$^\circ$.
\end{itemize}
The PFOV configuration allows us to test scenarios in which a participant may leave the camera field of view and still be heard. Notice that since ground-truth annotations are available for the full field of view, it is possible to assess the performance of the tracker using audio observations only, as well as to analyse the behavior of the tracker when it switches from audio-only tracking to audio-visual tracking.

\subsection{The AV16.3 Dataset}
\label{sec:av16.3}

\addnote[av163-description]{1}
{
We also used the twelve recordings of the AV16.3 dataset  \cite{lathoud2004av16} to evaluate the proposed method and to compare it with \cite{kilicc2015audio} and with \cite{kilicc2016mean}. The dataset was recorded in a meeting room. The videos were recorded at 25~FPS with three cameras fixed on the room ceiling. 
The image resolution is of $288\times360$ pixels. The audio signals were recorded with two eight-microphone circular arrays, both placed onto a table top, and 
sampled at $16$~kHz. In addition, the dataset comes with internal camera calibration parameters, as well as with external calibration parameters, namely 
camera-to-camera and microphone-array-to-camera calibration parameters.  We note that the scenarios associated with AV16.3 are somehow artificial in the sense 
that \textit{the participants speak simultaneously and continuously}. This stays in contrast with the AVDIAR recordings where people take speech turns in 
informal conversations. 
}

\subsection{Audio Features}
\label{subsec:audio_processing}
In the case of AVDIAR, the STFT (short-time Fourier transform) \cite{gold2011speech} is applied to each microphone signal using a 16~ms Hann window (256 audio samples per window) and with an 8~ms shift between successive windows (50\% overlap), leading to 128 frequency bins and to 125 audio FPS. Inter-microphone spectral features are then computed using \cite{li2017multiple}. These features -- referred to in \cite{li2017multiple} as \textit{direct-path relative transfer function (DP-RTF) features} -- are robust against background noise and against reverberations, hence they do not depend on the acoustic properties of the recording room, as they encode the direct path from the audio source to the microphones. \addnote[face-orientation]{2}{Nevertheless, they may depend on the orientation of the speaker's face. If the microphones are positioned behind a speaker, the direct-path sound wave (from the speaker to the microphones) propagates through the speaker's head, hence it is attenuated. This may have a negative impact on the direct-to-reverberation ratio. Here we assume that, altogether, this has a limited effect.}

The audio features are averaged over five audio frames in order to be properly aligned with the video frames. The feature vector is then split into $K=16$ sub-bands, each sub-band being composed of $J=8$ frequencies; sub-bands  with low energy are disregarded. This yields the set of audio observations at $t$, $\{\gvect_{tk}\}_{k=1}^{K_t}$, $K_t\leq K$ (see Section~\ref{subsec:audio-model} and Appendix~\ref{app:gllim}). 
\addnote[doa-for-others]{1}
{
Interestingly, the computed inter-microphone DP-RTF features can be mapped onto the image plane and hence they can be used to estimate directions of arrival (DOAs). Please consult
\cite{DeleforgeForbesHoraud2015,deleforge2015colocalization} for more details.}

\begin{figure}[t!]
\begin{center}
\includegraphics[width = 0.3\textwidth]{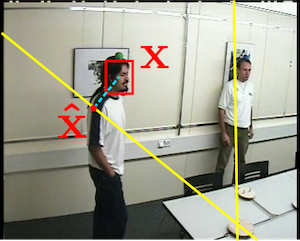}
\end{center}
\caption{\label{fig:doa-example} This figure displays two DOAs, associated with one microphone array (bottom left), projected onto the image plane, and illustrates the geometric relationship between a DOA and the current location of a speaker.}
\end{figure}

\addnote[DOA-features]{2}
{Alternatively, one can compute DOAs explicitly from time differences of arrival (TDOAs) between the microphones of a microphone array, provided that the inter-microphone geometry is known. The disadvantage is that DOAs based on TDOAs assume free-field acoustic-wave propagation and hence they don't have a built-in reverberation model.}
Moreover, if the camera parameters are known and if the camera location (extrinsic parameters) is known in the coordinate frame of the microphone array, as is the case with the AV16.3 dataset, it is possible to project DOAs onto the image plane.  We use the multiple-speaker DOA estimator of \cite{lathoud2005sector} as it provides accurate results for the AV16.3 sensor setup \cite{lathoud2004av16}.
Let $d_{tk}$ be the line corresponding to the projection of a DOA onto the image plane and let $\xvect_{tn}$ be the location of person $n$ at time $t$. It is straightforward to determine the point $\hat{\xvect}_{tk} \in d_{tk}$ the closest to $\xvect_{tn}$, e.g. Fig.~\ref{fig:doa-example}. Hence the inter-channel spectral features $\{\gvect_{tk}\}_{k=1}^{K_t}$ are replaced with $\{\hat{\xvect}_{tk}\}_{k=1}^{K_t}$ and \eqref{eq:affine-mapping} is replaced with:
\begin{align}
\label{eq:audio model av163}
p (  \hat{\xvect}_{tk} |&  \xvect_{tn},  B_{tk} =n)  =  \\
&
\begin{cases}
\mathcal{N} (\hat{\xvect}_{tk} ; \xvect_{tn}, \sigma \Imat) & \textrm{if} \: 1\leq n \leq N \\
\mathcal{U} (\hat{\xvect}_{tk} ; \textrm{vol} (\mathcal{X})) & \textrm{if} \: n=0,
\end{cases} \nonumber
\end{align}
where $\sigma \Imat$ is an isotropic covariance that models the uncertainty of the DOA, e.g. Fig.~\ref{fig:seq45-3p-1111}, third row.

\subsection{Visual Features}
\label{subsec:visual_processing}
In both AVDIAR and AV16.3 datasets participants do not always face the cameras and hence face detection is not robust. Instead we use the person detector 
of~\cite{cao2017realtime} from which we infer a body bounding-box and a head bounding-box. We use the person re-identification CNN-based 
method~\cite{zheng2017person} to extract an embedding (i.e. a person descriptor) from the body bounding-box. This yields the feature vectors 
$\{\uvect_{tm}\}_{m=1}^{M_t}\subset\mathbb{R}^{2048}$ (Section~\ref{subsec:visual-model}). Similarly, the center, width and height of the head bounding-box 
yield the observations $\{\vvect_{tm}\}_{m=1}^{M_t}\subset\mathbb{R}^{4}$ at each frame $t$.

\subsection{Evaluation Metrics}

We used standard multi-object tracking (MOT) metrics~\cite{milan2016mot16} to quantitatively evaluate the performance of the proposed tracking algorithm. The multi-object tracking 
accuracy (MOTA) is the most commonly used metric for MOT. It is a combination of false positives (FP), false negatives (FN; i.e.\ missed persons), and identity 
switches (IDs), and is defined as:  
\begin{equation}
	\textrm{MOTA} = 100 \left( 1 - \frac{\sum_{t} (\textrm{FP}_{t} + \textrm{FN}_{t} + \textrm{IDs}_{t})}{\sum_{t} \textrm{GT}_{t}} \right),
\end{equation}
where GT stands for the ground-truth person trajectories. After comparison with GT trajectories, each estimated trajectory can be classified as mostly tracked 
(MT) and mostly lost (ML) depending on whether a trajectory is covered by correct estimates more than $80\%$ of the time (MT) or less than $20\%$ of the time 
(ML). In the tables below, MT and ML indicated the percentage of ground-truth tracks under each situation.


\addnote[ospa-t]{1}{
In addition to MOT, we also used the OSPA-T metric~\cite{ristic2011metric}. OSPA-T is based on a distance between two point sets and combines various aspects 
of tracking performance, such as timeliness, track accuracy, continuity, data associations and false tracks. It should be noted that OSPA-T involves a number 
of parameters whose values must be provided in advance. We used the publicly available code provided by one of the authors of~\cite{ristic2011metric} for 
computing the OSPA-T scores in all the experimental evaluations reported below.\footnote{\url{http://ba-tuong.vo-au.com/codes.html}}
}

In our experiments, the threshold of overlap to consider that a ground truth is covered by an estimation is set to $0.1$ intersection over union (IoU).
In the PFOV configuration, we need to evaluate the audio-only tracking, i.e. the speakers are in the blind areas.   
As mentioned before, audio localization is less accurate than visual localization. 
Therefore, for evaluating the audio-only tracker we relax by a factor of two the expected localization accuracy with respect to the audio-visual localization accuracy. 


\subsection{Benchmarking with Baseline Methods}

\begin{small}
\begin{table}[t!]
	\centering
	\caption{OSPA-T and MOT scores for the living-room sequences (full camera field of view)
	} 
	\resizebox{.5\textwidth}{!}{
	\begin{tabular}{c|c|c|cc|c|cc}
	\toprule
	Method   &OSPA-T($\downarrow$) & MOTA($\uparrow$)  & FP($\downarrow$) & FN($\downarrow$) & IDs($\downarrow$) &MT($\uparrow$)     & ML($\downarrow$)\\  
	\midrule 
\cite{kilicc2015audio} &28.12 & 10.37  & 44.64 \% & 43.95\% & 732   & 20\%  & 7.5 \%   \\
\midrule  

\cite{kilicc2016mean} & 30.03 & 18.96   & 8.13 \%& 72.09\% & 581  & 17.5\%  & 52.5\% \\ 
\midrule   
 \cite{ba2016line} & \textbf{14.79} & \textbf{96.32}  & \textbf{1.77\%} & \textbf{1.79\%} & \textbf{80}  & \textbf{92.5\%} & \textbf{0\%}  \\ 
\midrule 
VAVIT   &17.05 & 96.03 & 1.85\% & 2.0\% & 86   & \textbf{92.5\%} & \textbf{0\%} \\ 
	\bottomrule
	\end{tabular}\label{tab::livingRoom_FULL}
	}\vspace{-2mm}
\end{table} 
\end{small}

\begin{small}
\begin{table}[t!]
	\centering
	\caption{OSPA-T and MOT scores for the meeting-room sequences (full camera field of view).} 
	\resizebox{.5\textwidth}{!}{
	\begin{tabular}{c|c|c|cc|c|cc}
	\toprule
	Method &OSPA-T ($\downarrow$)& MOTA($\uparrow$)  & FP($\downarrow$) & FN($\downarrow$) & IDs($\downarrow$) & MT($\uparrow$) & ML($\downarrow$)  \\ 
	\midrule
\cite{kilicc2015audio} &5.76 & 62.43  & 18.63\%  & 17.19\% & 297  & 70.59 \% & \textbf{0\%}  \\ 
\midrule  	

\cite{kilicc2016mean} 	& 7.83 &28.48  & 0.93\% & 69.68\% & 155 & 0 \% & 52.94\% \\  
 \midrule 

\cite{ba2016line} &\textbf{3.02} & \textbf{98.50} & \textbf{0.25\%} & \textbf{1.11\%} & \textbf{25}   & \textbf{100.00\%}  & \textbf{0\%} \\   
\midrule
VAVIT   &3.57& 98.16  & 0.38\% & 1.27\% & 32  & \textbf{100.00\%}  & \textbf{0\%}   \\ 
	\bottomrule
	\end{tabular}\label{tab::meeting-room_FULL}
	}\vspace{-2mm}
\end{table} 
\end{small}

\begin{small}
\begin{table}[t!]
	\centering
	\caption{OSPA-T and MOT scores for the living-room sequences (partial camera field of view).} 
	\resizebox{.5\textwidth}{!}{
	\begin{tabular}{c|c|c|cc|c|cc}
	\toprule
	Method  &OSPA-T($\downarrow$)  & MOTA($\uparrow$) & FP($\downarrow$) & FN($\downarrow$) & IDs($\downarrow$) & MT($\uparrow$)  & ML($\downarrow$) \\ 
	\midrule 
\cite{kilicc2015audio}  & 28.14 & 17.82  & 36.86\% & 42.88\% & 1722  & 32.50\% & 7.5\% \\
\midrule

 \cite{kilicc2016mean} &  29.73 & 20.61   & 5.54\% & 72.45\% & 989   & 12.5\%  & 40\%\\ 
 \midrule 

\cite{ba2016line}  & 22.25 & 66.39 & \textbf{0.48\%} & 32.95\% & \textbf{129}    & 45\%  & 7.5\%  \\
\midrule
VAVIT   &\textbf{21.77} & \textbf{69.62} & 8.97\% & \textbf{21.18\%} & 152  & \textbf{70\%}  & \textbf{5\%}\\
	\bottomrule
	\end{tabular}\label{tab::livingRoom_PART}
	}\vspace{-2mm}
\end{table} 
\end{small}

\begin{small}
\begin{table}[t!]
	\centering
	\caption{OSPA-T and  MOT scores for the meeting-room sequences (partial camera field of view).} 
	\resizebox{.5\textwidth}{!}{
	\begin{tabular}{c|c|c|cc|c|cc}
	\toprule
	Method  & OSPA-T($\downarrow$) & MOTA($\uparrow$)  & FP($\downarrow$) & FN($\downarrow$) & IDs($\downarrow$) 
	& MT($\uparrow$) & ML($\downarrow$) \\ 
	\midrule 
\cite{kilicc2015audio} &7.23 & 29.04  & 23.05\% & 45.19 \% & 461  & 29.41\%  & 17.65\%   \\ 
\midrule  

 \cite{kilicc2016mean}  & 8.17  & 26.95  & 1.05\% & 70.62\% & 234 & 5.88\%  & 52.94\%  \\
\midrule 
\cite{ba2016line} & \textbf{5.80} & 64.24 & \textbf{0.43\%} & 35.18\% & \textbf{24}    & 36.84\%  & 15.79\%  \\ 
\midrule
VAVIT   & 5.81 & \textbf{65.27}  & 5.07\% & \textbf{29.5\%} & 26   & \textbf{47.37\%}  & \textbf{10.53\%} \\
	\bottomrule
	\end{tabular}\label{tab::meeting-room_PART}
	}\vspace{-2mm}
\end{table} 
\end{small}

\begin{small}
\begin{table}[t!]
	\centering
	\caption{OSPA-T and MOT scores obtained with the AV16.3 dataset.} 
	\resizebox{.5\textwidth}{!}{
	\begin{tabular}{c|c|c|cc|c|cc}
	\toprule
	Method   & OSPA-T ($\downarrow$) & MOTA($\uparrow$) & FP($\downarrow$) & FN($\downarrow$) & IDs($\downarrow$) 
	& MT($\uparrow$) & ML($\downarrow$) \\ 
\midrule  

\cite{kilicc2016mean} &  17.28 & 36.4   & 16.72\% & 42.22\% & 765  & 11.11\%  & \textbf{0\%} \\
\midrule 
\cite{ba2016line} & 13.32 & 82.9   & \textbf{5.29\%} & 11.5 \% & 51    & 85.2\%  & \textbf{0\%} \\ 
\midrule
VAVIT   & \textbf{10.88} & \textbf{84.1}  & 6.51\% & \textbf{9.18\%} & \textbf{29}  & \textbf{92.6\%}  & \textbf{0\%} \\ 

	\bottomrule
	\end{tabular}\label{tab::AV16.3}
	}\vspace{-5mm}
\end{table} 
\end{small}

To quantitatively evaluate its performance, we benchmarked the proposed method with two state-of-the-art audio-visual tracking methods. The first one is the audio-assisted video adaptive particle filtering (AS-VA-PF) method of \cite{kilicc2015audio}, and the second one is the sparse audio-visual mean-shift sequential Monte-Carlo probability hypothesis density (AV-MSSMC-PHD) method of \cite{kilicc2016mean}.
\addnote[visual-observations-2]{2}
{
Notice that both these methods do not make recourse to a person detector as they use a tracking-by-detection paradigm. This stays in contrast with our method which uses a person detector and probabilistically assigns each detection to each person. In principle, the baseline methods can be modified to accept person detection as visual information. However, we did not modify the baseline methods and used the software provided by the authors of \cite{kilicc2015audio} and \cite{kilicc2016mean}.
}
Sound locations are used to reshape the typical Gaussian noise distribution of particles in a propagation step, then \cite{kilicc2015audio} uses the particles to weight the observation model. \cite{kilicc2016mean} uses audio information to improve the performance and robustness of a visual SMC-PHD filter. Both \cite{kilicc2015audio} and \cite{kilicc2016mean} require input from a multiple sound-source localization (SSL) algorithm. 
\addnote[audio-observations-datasets]{1}{
In the case of AVDIAR recordings, the multi-speaker localization method proposed in \cite{li2017multiple} is used to provide input to \cite{kilicc2015audio} and \cite{kilicc2016mean}.\footnote{The authors of \cite{kilicc2015audio} and \cite{kilicc2016mean} kindly provided their software packages.} In the case of AV16.3 recordings the method of  \cite{lathoud2004av16} is used to provide DOAs to \cite{kilicc2015audio,kilicc2016mean} and to our method, as explained in Section~\ref{subsec:audio_processing} above.
}

We also compare the proposed method with a visual multiple-person tracker, more specifically the \textit{online Bayesian variational tracker} (OBVT) of \cite{ba2016line}, which is based on a similar variational inference as the one presented in this paper. In \cite{ba2016line} visual observations were provided by color histograms.
In our benchmark,
for the sake of fairness, the proposed tracker and \cite{ba2016line} share the same visual observations, as described in Section~\ref{subsec:visual_processing}.

The OSPA-T and MOT scores obtained with these methods as well as the proposed method are reported in Table~\ref{tab::livingRoom_FULL}, Table~\ref{tab::meeting-room_FULL}, Table~\ref{tab::livingRoom_PART}, Table~\ref{tab::meeting-room_PART}, and Table~\ref{tab::AV16.3}. The symbols $\uparrow$ and $\downarrow$ indicate higher the better and  lower the better, respectively. In the case of AVDIAR, we report results with both meeting-room and living-room in the two configurations: FFOV, Table~\ref{tab::livingRoom_FULL} and Table~\ref{tab::meeting-room_FULL} and PFOV, Table~\ref{tab::livingRoom_PART} and Table~\ref{tab::meeting-room_PART}. In the case of AV16.3 we report results with the twelve recordings commonly used by audio-visual tracking algorithms, Table~\ref{tab::AV16.3}.

The most informative metrics are OSPA-T and MOTA (MOT accuracy) and one can easily see that both \cite{ba2016line} and the proposed method outperform the other two methods. 
The poorer performance of both \cite{kilicc2015audio} and \cite{kilicc2016mean} for all the configurations is generally explained by the fact that these two methods expect audio and visual observations to be simultaneously available. In particular, \cite{kilicc2015audio} is not robust against visual occlusions, which leads to poor IDs (identity switches) scores. 

The AV-MSSMC-PHD method \cite{kilicc2016mean} uses audio information in order to count the number of speakers. 
\addnote[FN-ID]{1}{
In practice, we noticed that the algorithm behaves differently with the two datasets. In the case of AVDIAR, 
we noticed that the algorithm assigns several visible participants to the same audio source, since in most of the cases there is only one active audio source at a time.  In the case of AV16.3 the algorithm performs much better, since participants speak simultaneously and continuously.
This explains why both FN (false negatives) and IDs (identity switches) scores are high in the case of AVDIAR,
i.e. Tables~\ref{tab::livingRoom_FULL}, \ref{tab::meeting-room_FULL}, and \ref{tab::livingRoom_PART}.
}

One can notice that in the case of FFOV, \cite{ba2016line} and the proposed method yield similar results in terms of OSPA-T and MOT scores: both methods exhibit 
low OSPA-T, FP, FN and IDs scores and, consequently, high MOTA scores. Moreover, they have very good MT and ML scores (out of 40 sequences 37 are mostly 
tracked, 3 are partially tracked, and none is mostly lost). As expected, the inferred trajectories are more accurate for visual tracking (whenever visual 
observations are available) than for audio-visual tracking: indeed, the latter fuses visual and audio observations which slightly degrades the accuracy because 
audio localization is less accurate than visual localization. 

As for the PFOV configuration (Table~\ref{tab::livingRoom_PART} and Table~\ref{tab::meeting-room_PART}), the proposed algorithm yields the best MOTA scores both 
for meeting-room and for living-room. Both \cite{kilicc2015audio} and \cite{kilicc2016mean} have difficulties when visual information is not available: both 
these algorithms fail to track speakers when they walk outside the visual field of view. While \cite{kilicc2016mean} can detect a speaker when it re-enters the 
visual field of view, \cite{kilicc2015audio} cannot. Obviously, the visual-only tracker \cite{ba2016line} fails outside the camera field of view.

\begin{figure*}[p!]
\begin{center}
\begin{tabular}{cccc}
\hspace{-2ex}
\includegraphics[width = 0.23\textwidth]{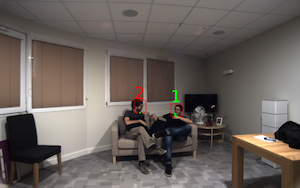}&
\hspace{-2ex}
\includegraphics[width = 0.23\textwidth]{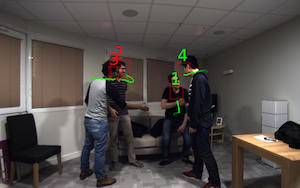}&
\hspace{-2ex}
\includegraphics[width = 0.23\textwidth]{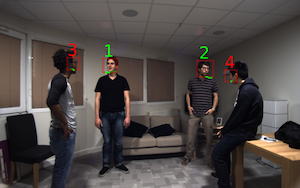}&
\hspace{-2ex}
\includegraphics[width = 0.23\textwidth]{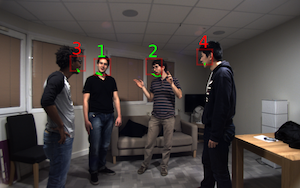}
\\
\hspace{-2ex}
\includegraphics[width = 0.23\textwidth]{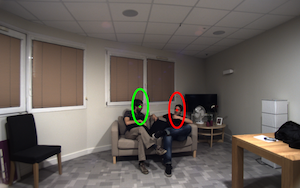}&
\hspace{-2ex}
\includegraphics[width = 0.23\textwidth]{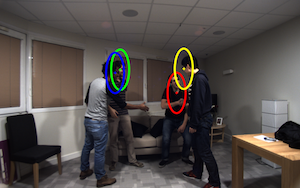}&
\hspace{-2ex}
\includegraphics[width = 0.23\textwidth]{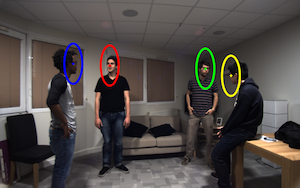}&
\hspace{-2ex}
\includegraphics[width = 0.23\textwidth]{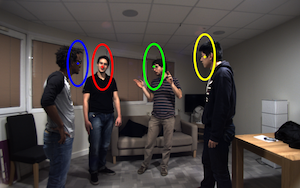}
\\
\hspace{-2ex}
\includegraphics[width = 0.23\textwidth]{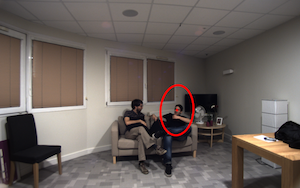}&
\hspace{-2ex}
\includegraphics[width = 0.23\textwidth]{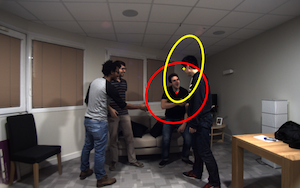}&
\hspace{-2ex}
\includegraphics[width = 0.23\textwidth]{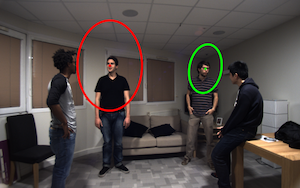}&
\hspace{-2ex}
\includegraphics[width = 0.23\textwidth]{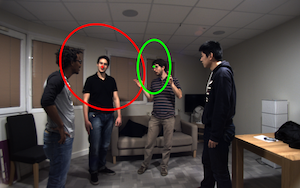}
\\
\hspace{-2ex}
\includegraphics[width = 0.23\textwidth]{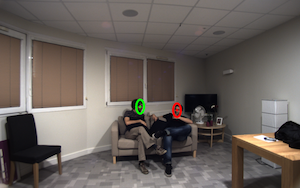}&
\hspace{-2ex}
\includegraphics[width = 0.23\textwidth]{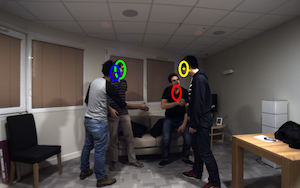}&
\hspace{-2ex}
\includegraphics[width = 0.23\textwidth]{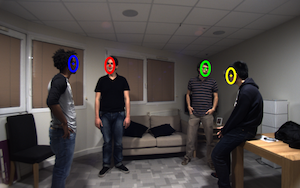}&
\hspace{-2ex}
\includegraphics[width = 0.23\textwidth]{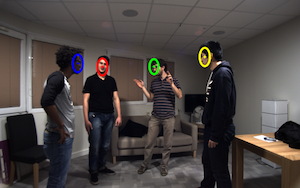}
\end{tabular}
\end{center}
\caption{Four frames sampled from Seq13-4P-S2M1 (living room). First row: green digits denote speakers while red digits denote silent participants. Second, third and fourth rows:  the ellipses visualize the visual, audio, and dynamic covariances, respectively, of each tracked person. The tracked persons are color-coded: green, yellow, blue, and red.
}
\label{fig:Seq13-4P-S2M1}
\end{figure*} 

\begin{figure*}[p!]
\begin{center}
\begin{tabular}{cccc}
\hspace{-2ex}
\includegraphics[width = 0.23\textwidth]{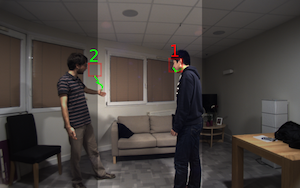}&
\hspace{-2ex}
\includegraphics[width = 0.23\textwidth]{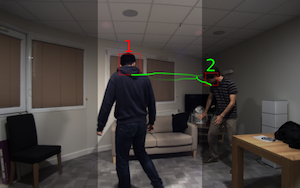}&
\hspace{-2ex}
\includegraphics[width = 0.23\textwidth]{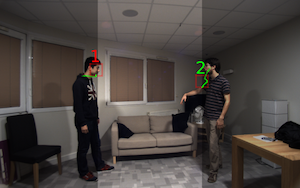}&
\hspace{-2ex}
\includegraphics[width = 0.23\textwidth]{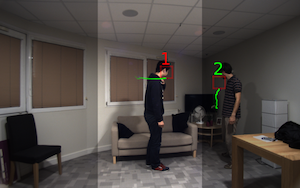}
\\
\hspace{-2ex}
\includegraphics[width = 0.23\textwidth]{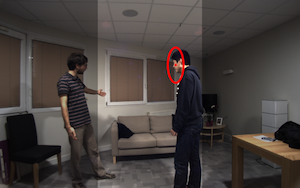}&
\hspace{-2ex}
\includegraphics[width = 0.23\textwidth]{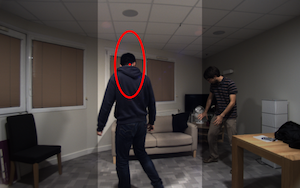}&
\hspace{-2ex}
\includegraphics[width = 0.23\textwidth]{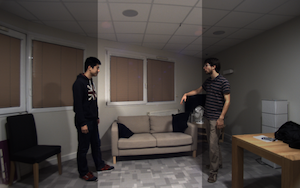}&
\hspace{-2ex}
\includegraphics[width = 0.23\textwidth]{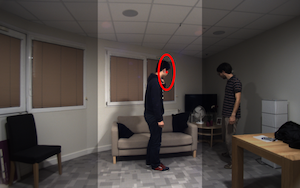}
\\
\hspace{-2ex}
\includegraphics[width = 0.23\textwidth]{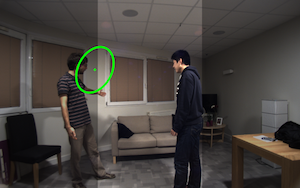}&
\hspace{-2ex}
\includegraphics[width = 0.23\textwidth]{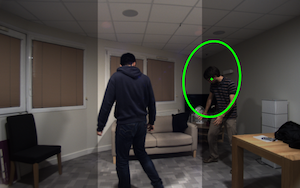}&
\hspace{-2ex}
\includegraphics[width = 0.23\textwidth]{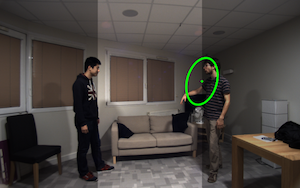}&
\hspace{-2ex}
\includegraphics[width = 0.23\textwidth]{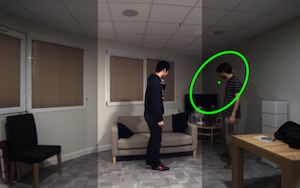}
\\
\hspace{-2ex}
\includegraphics[width = 0.23\textwidth]{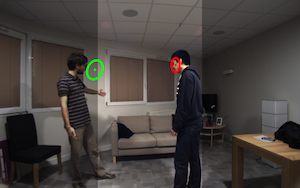}&
\hspace{-2ex}
\includegraphics[width = 0.23\textwidth]{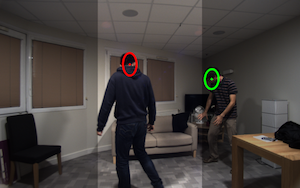}&
\hspace{-2ex}
\includegraphics[width = 0.23\textwidth]{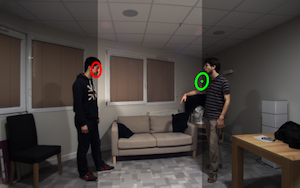}&
\hspace{-2ex}
\includegraphics[width = 0.23\textwidth]{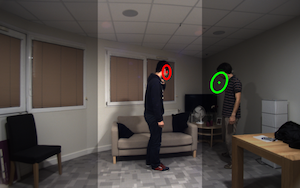}
\end{tabular}
\end{center}
\caption{Four frames sampled from Seq19-2P-S1M1 (living room). The camera field of view is limited to the central strip. Whenever the participants are outside the central strip, the tracker entirely relies on audio observations and on the model's dynamics.}
\label{fig:Seq19-2P-S1M1}
\end{figure*} 

\begin{figure*}[t!]
\begin{center}
\begin{tabular}{cccc}
\hspace{-2ex}
\includegraphics[width = 0.20\textwidth]{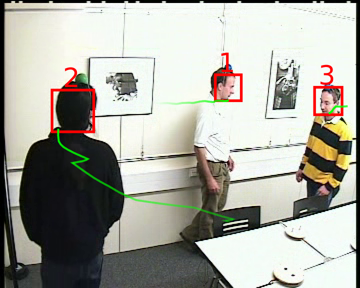}&
\hspace{-2ex}
\includegraphics[width = 0.20\textwidth]{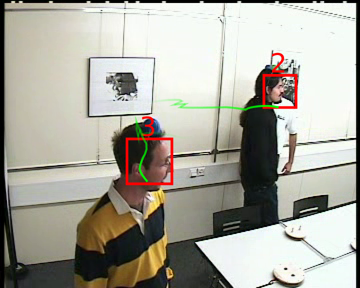}&
\hspace{-2ex}
\includegraphics[width = 0.20\textwidth]{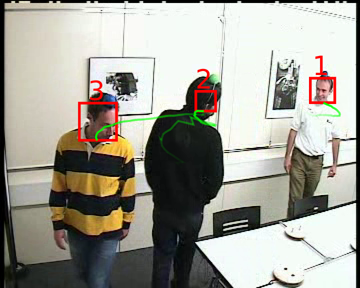}&
\hspace{-2ex}
\includegraphics[width = 0.20\textwidth]{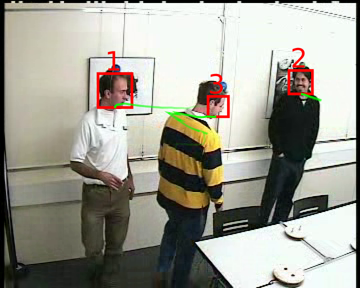}
\\
\hspace{-2ex}
\includegraphics[width = 0.20\textwidth]{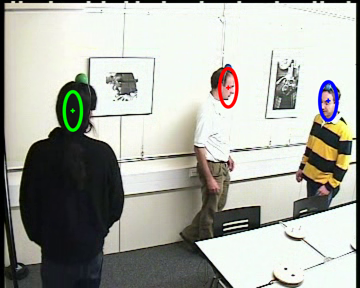}&
\hspace{-2ex}
\includegraphics[width = 0.20\textwidth]{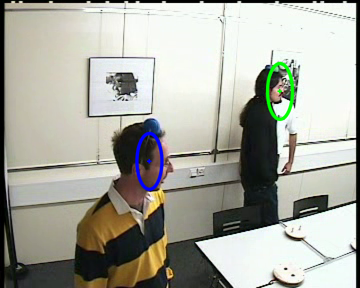}&
\hspace{-2ex}
\includegraphics[width = 0.20\textwidth]{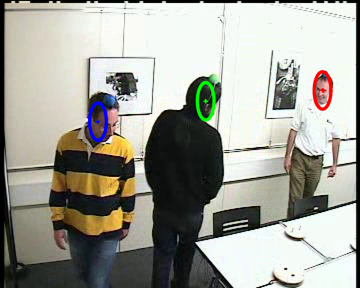}&
\hspace{-2ex}
\includegraphics[width = 0.20\textwidth]{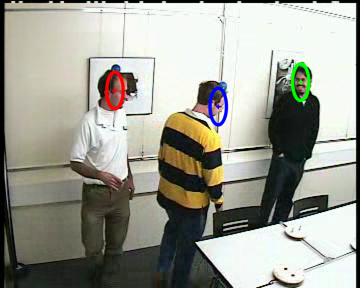}
\\
\hspace{-2ex}
\includegraphics[width = 0.20\textwidth]{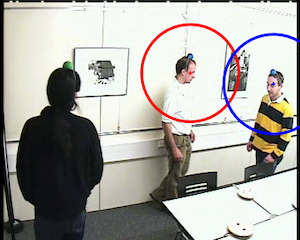}&
\hspace{-2ex}
\includegraphics[width = 0.20\textwidth]{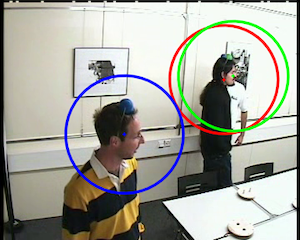}&
\hspace{-2ex}
\includegraphics[width = 0.20\textwidth]{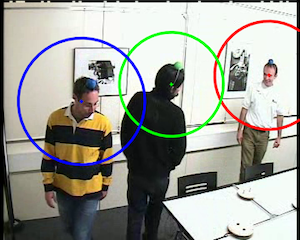}&
\hspace{-2ex}
\includegraphics[width = 0.20\textwidth]{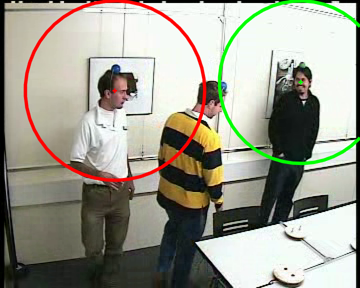}
\\
\hspace{-2ex}
\includegraphics[width = 0.20\textwidth]{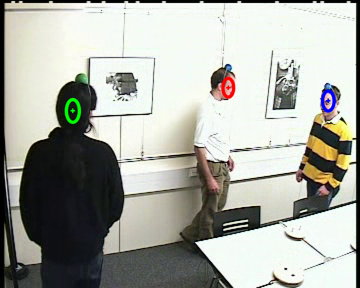}&
\hspace{-2ex}
\includegraphics[width = 0.20\textwidth]{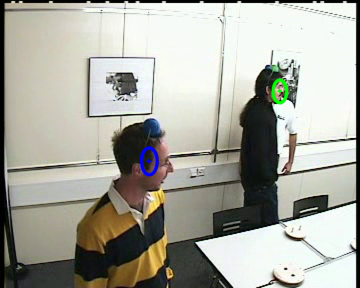}&
\hspace{-2ex}
\includegraphics[width = 0.20\textwidth]{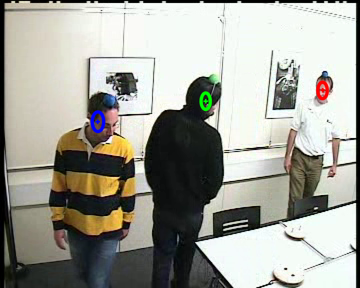}&
\hspace{-2ex}
\includegraphics[width = 0.20\textwidth]{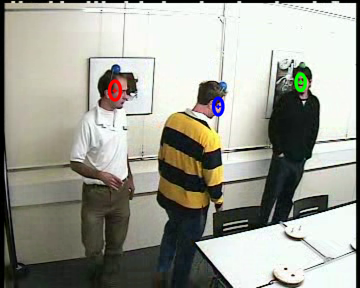}
\end{tabular}
\end{center}
\caption{Four frames sampled from  seq45-3p-1111 of AV16.3. In this dataset, the participants speak simultaneously and continuously. }
\label{fig:seq45-3p-1111}
\end{figure*} 

\begin{figure*}[t!b!]
\begin{center}
\begin{tabular}{cccc}
\includegraphics[width = 0.225\textwidth]{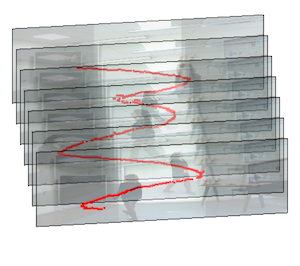}&
\includegraphics[width = 0.225\textwidth]{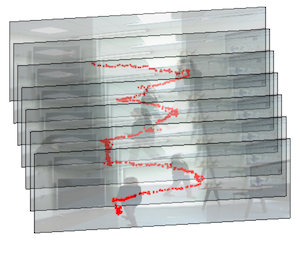} &
\includegraphics[width = 0.225\textwidth]{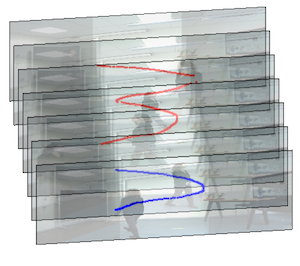} &
\includegraphics[width = 0.225\textwidth]{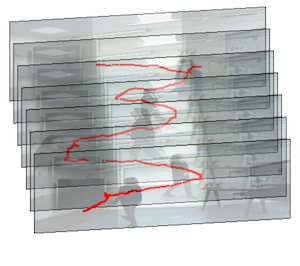}\\
{\small (a) Ground-truth trajectory} & {\small (b) AS-VA-PF \cite{kilicc2015audio}} & {\small (c) OBVT \cite{ba2016line}} & {\small  (d) VAVIT (proposed)}
\end{tabular}
\end{center}
\caption{Trajectories associated with a tracked person under the PFOV configuration (sequence Seq22-1P-S0M1 recorded in meeting room). The ground-truth trajectory (a) corresponds to the center of the bounding-box of the head. The trajectory  (b) obtained with \cite{kilicc2015audio} is non-smooth. Both \cite{kilicc2015audio} and \cite{ba2016line} fail to track outside the camera field of view. In the case of the OBVT trajectory (c), there is an identity switch, from ``red" (before the person leaves the visual field of view) to ``blue" (after the person re-enters in the visual field of view). }
\label{fig:Seq22-1P-S0M1}
\end{figure*}

\subsection{Audio-Visual Tracking Examples}

We now provide and discuss results obtained with three AVDIAR recordings and one AV16.3 recording, namely the FFOV recording Seq13-4P-S2-M1 (Fig.~\ref{fig:Seq13-4P-S2M1}), the PFOV recordings Seq19-2P-S1M1 (Fig.~\ref{fig:Seq19-2P-S1M1}) and Seq22-1P-S0M1 (Fig.~\ref{fig:Seq22-1P-S0M1}), and the seq45-3p-1111 recording of AV16.3 (Fig.~\ref{fig:seq45-3p-1111}).\footnote{\url{https://team.inria.fr/perception/research/variational_av_tracking/}} All these recordings are challenging in terms of audio-visual tracking: participants are seated, then they stand up or they wander around. In the case of AVDIAR, some participants take speech turns and interrupt each other, while others remain silent. 

The first rows of Fig.~\ref{fig:Seq13-4P-S2M1}, Fig.~\ref{fig:Seq19-2P-S1M1} and  Fig.~\ref{fig:seq45-3p-1111} show four frames sampled from two AVDIAR recordings and one AV16.3 recording, respectively. The second rows show ellipses of constant density that correspond to visual uncertainty (covariances). 
The third rows show the audio uncertainty. The audio uncertainties (covariances) are much larger than the visual ones since audio localization is less accurate 
than visual localization. The fourth rows shows the contribution of the dynamic model to the uncertainty, i.e. the inverse of the precision (\#3) in eq.  
\eqref{eq:posteriordistribution-of-Xtn-cov}. Notice that these ``dynamic" covariances are small, in comparison with the ``observation" covariances. This 
ensures tracking continuity (smooth tracjectories) when audio or visual observations are either weak or totally absent.
Fig.~\ref{fig:Seq19-2P-S1M1} shows a tracking example with a partial camera field of view (PFOV) configuration. In this case, audio and visual observations are barely available simultaneously. The independence of the visual and audio observation models and their fusion within the same dynamic model guarantees robust tracking in this case.

Fig.~\ref{fig:Seq22-1P-S0M1} shows the ground-truth trajectory of a person and the trajectories estimated with the audio-visual tracker \cite{kilicc2015audio}, with the visual tracker  \cite{ba2016line}, and with the proposed method. The ground-truth trajectory corresponds to a sequence of bounding-box centers. Both \cite{kilicc2015audio} and \cite{ba2016line} failed to estimate a correct trajectory. Indeed, \cite{kilicc2015audio} requires simultaneous availability of audio-visual data while  \cite{ba2016line} cannot track outside the visual field of view. Notice the non-smooth trajectory obtained with \cite{kilicc2015audio} in comparison with the smooth trajectories obtained with variational inference, i.e. \cite{ba2016line} and proposed.

\subsection{Computation Times}
\label{sec:Execution time} 
\addnote[exec-time]{1}{
Matlab implementations of algorithms \cite{kilicc2015audio}, \cite{kilicc2016mean},  \cite{ba2016line} and VAVIT were run on an Intel(R) 8-core 2.40~GHz CPU E5-2609 equipped with 32~GB of RAM and with a GeForce GTX 1070 GPU. The computation times provided in Table~\ref{tab:time-avdiar} correspond to the total number of frames associated with all the sequences available in the two datasets. Both \cite{ba2016line} and VAVIT necessitate a person detector. The CNN-based person detector runs on the same computer at 2~FPS. Person detection is run offline. 
}

\subsection{Speaker Diarization Results}
\label{sec:diarization results} 
As already mentioned in Section~\ref{sec:diarization}, speaker diarization information can be extracted from the output of the proposed VAVIT algorithm. 
Notice that, while audio diarization is an extremely well investigated topic, audio-visual diarization has received much less attention. In \cite{noulas2012multimodal} it is proposed an audio-visual diarization method based on a dynamic Bayesian network that is applied to video conferencing. Their method assumes that participants take speech turns with a small silent interval between turns, which is an unrealistic hypothesis in the general case. 
The diarization method of \cite{minotto2015multimodal} requires audio, depth and RGB data. 
More recently, \cite{gebru2018a} proposed a Bayesian dynamic model for audio-visual diarization that takes as input fused audio-visual information. Since diarization is not the main objective of this paper, we only compared our diarization results with \cite{gebru2018a}, which achieves state of the art results, and with the diarization toolkit of \cite{vijayasenan2012diartk} which only considers audio information.

The diarization error rate (DER) is generally used as a quantitative measure. As is the case with MOT, DER combines false positives (FP), false negatives (FN) and identity swithches (IDs) scores within a single metric.
The NIST-RT evaluation toolbox\footnote{\url{https://www.nist.gov/itl/iad/mig/rich-transcription-evaluation}} is used. The results obtained with  \cite{vijayasenan2012diartk,gebru2018a} and with the proposed method are reported in Table~\ref{tab:DIAR-FULL}, for both the full field-of-view and partial field-of-view configurations (FFOV and PFOV). The proposed method performs better than the audio-only baseline method \cite{vijayasenan2012diartk}. In comparison with \cite{gebru2018a}, the proposed method performs slightly less well despite the lack of a special-purpose diarization model. Indeed, \cite{gebru2018a} implements diarization within a hidden Markov model (HMM) that takes into account both diarization dynamics and the audio activity observed at each time step, whereas our method is only based on observing the audio activity over time.

\begin{figure*}[t!]
\centering
\includegraphics[width = 0.9\textwidth]{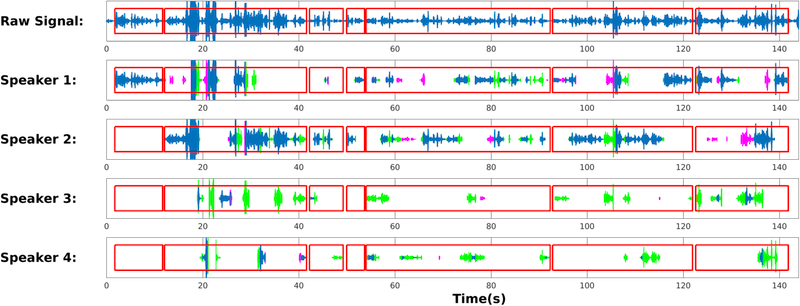}
\caption{\label{fig:Seq13-4P-S2M1-diar} Diarization results obtained with Seq13-4P-S2M1 (FFOV). The first row shows the audio signal recorded with one of the microphones. The red boxes show the result of the voice activity detector which is applied to all the microphone signals prior to tracking. For each speaker, correct detections are shown in blue, missed detections are shown in green, and false positives are shown in magenta}
\end{figure*}

\begin{figure*}[t!]
\begin{center}
\includegraphics[width = 0.8\textwidth]{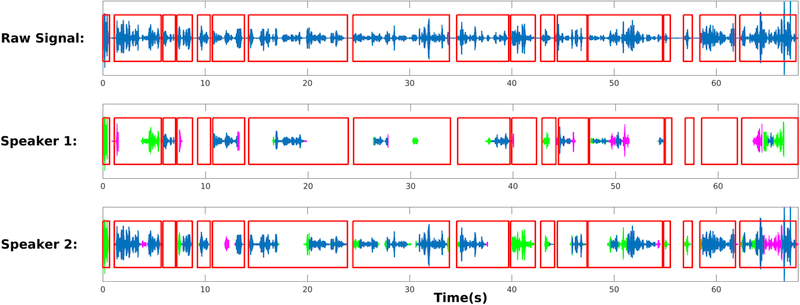}
\end{center}
\caption{\label{fig:Seq19-2P-S1M1-diar} Diarization results obtained with Seq19-2P-S1M1 (PFOV).}
\end{figure*}

The ability of the proposed audio-visual tracker to perform diarization is illustrated in Fig.~\ref{fig:Seq13-4P-S2M1-diar} and in Fig.~~\ref{fig:Seq19-2P-S1M1-diar}
with a FFOV sequence (Seq13-4P-S2-M1, Fig.~\ref{fig:Seq13-4P-S2M1}) and with a PFOV sequence (Seq19-2P-S1M1, Fig.~\ref{fig:Seq19-2P-S1M1}), respectively.

\begin{table}[t!]
	\centering
	\caption{\label{tab:time-avdiar} Computation times (in seconds). All four algorithms are implemented in Matlab and run on the same computer.} 
	\resizebox{.5\textwidth}{!}{
	\begin{tabular}{cccc}
	\toprule
	Methods & AVDIAR: Living room  & AVDIAR: Meeting room  & AV16.3\\
		\midrule 
		Number of frames & 26927 & 6031 & 11135\\
		
		\midrule 
\cite{kilicc2015audio} & 20821 &2424& - \\
\midrule  
 
\cite{kilicc2016mean} &10510&2267&611\\ 
\midrule   
 \cite{ba2016line} & \textbf{542} & \textbf{130} & \textbf{236}  \\ 
\midrule 
VAVIT  &3456&759&260\\ 
\bottomrule
	\end{tabular}	
	}
\end{table}

\begin{small}
\begin{table}[t!]
	\centering
	\caption{DER (diarization error rate) scores obtained with the AVDIAR dataset.} 
	\resizebox{.45\textwidth}{!}{
	\begin{tabular}{lcccc}
	\toprule
	Sequence & DiarTK \cite{vijayasenan2012diartk} & \cite{gebru2018a} & Proposed (FFOV) & Proposed (PFOV) \\
	\midrule
Seq01-1P-S0M1  & 43.19 & 3.32 &     1.64 & 1.86 \\
Seq02-1P-S0M1  & 49.9   & -        &  2.38  & 2.09 \\
Seq03-1P-S0M1  & 47.25  & -       &   6.59  & 14.65\\
Seq04-1P-S0M1  & 32.62& 9.44      &  4.96 & 10.45\\
Seq05-2P-S1M0  & 37.76& - &   29.76  & 30.78\\
Seq06-2P-S1M0  & 56.12&- &  14.72  & 15.83\\
Seq07-2P-S1M0  & 41.43&-&   42.36  & 37.56\\
Seq08-3P-S1M1  & 31.5&-&  38.4  & 48.86\\
Seq09-3P-S1M1  & 52.74&-&   38.26  & 68.81\\
Seq10-3P-S1M1  & 56.95&-&   54.26  & 54.04\\
Seq12-3P-S1M1  & 63.67&17.32&   44.67  & 47.25\\
Seq13-4P-S2M1  & 47.56&29.62&   43.45  & 43.17\\
Seq15-4P-S2M1  & 62.53&-&    41.49  & 64.38\\
Seq17-2P-S1M1  & 17.24&-&  16.53  & 15.63\\
Seq18-2P-S1M1  & 35.05&-&  19.55  & 20.58\\
Seq19-2P-S1M1  & 38.96&-&   26.47  & 27.84\\
Seq20-2P-S1M1  & 43.58&35.46&  38.24 & 44.3\\
Seq21-2P-S1M1  & 32.22 &20.93&   25.87 & 25.9\\
Seq22-1P-S0M1  & 23.53 &4.93&  2.79   & 3.32\\
Seq27-3P-S1M1  & 46.05 &18.72&  47.07  & 54.75\\
Seq28-3P-S1M1  & 30.68 &-&  23.54  & 31.77\\
Seq29-3P-S1M0  & 38.68 &-&  30.74  & 35.92\\
Seq30-3P-S1M1  & 51.15 &-&  49.71  & 57.94\\
Seq32-4P-S1M1  & 41.51 &30.20&  46.25  & 43.03\\
\midrule  
Overall        &42.58 &\textbf{18.88} &    28.73 &33.36 \\
	\bottomrule
	\label{tab:DIAR-FULL}
	\end{tabular}	
	}
\end{table}
\end{small}

\section{Conclusions}
\label{sec:conclusions}
We addressed the problem of tracking multiple speakers using audio and visual data. It is well known that the generalization of single-person tracking to multiple-person tracking is computationally intractable and a number of methods were proposed in the past. Among these methods, sampling methods based on particle filtering (PF) or on PHD filters have recently achieved the best tracking results. However, these methods have several drawbacks: (i)~the quality of the approximation of the filtering distribution increases with the number of particles, which also increases the computational burden, (ii)~the observation-to-person association problem is not explicitly modeled and a post-processing association mechanism must be invoked, and (iii) audio and visual observations must be available simultaneously and continuously. Some of these limitations were recently addressed both in \cite{kilicc2015audio} and in \cite{kilicc2016mean}, where audio observations were used to compensate the temporal absence of visual observations. Nevertheless, people speak with pauses and hence audio observations are rarely continuously available.

In contrast, we proposed a variational approximation of the filtering distribution and we derived a closed-form variational expectation-maximization algorithm. The observation-to-person association problem is fully integrated in our model, rather than as a post-processing stage.
The proposed VAVIT algorithm is able to deal with intermittent audio or visual observations, such that one modality can compensate the other modality, whenever one of them is noisy, too weak or totally missing. Using the OSPA-T and MOT scores we showed that the proposed method outperforms the PF-based method \cite{kilicc2015audio}.

\appendices
\section{An Audio Generative Model}
\label{app:gllim}
In this appendix we describe the audio observation model used in this paper. More precisely, we make explicit the generative model introduced in Section~
III-D, i.e. equation~(13).
For that purpose we consider a training set of audio features, or inter-channel spectral features (which in practice correspond to the real and imaginary parts of complex-valued Fourier coefficients) and their associated source locations, $\mathcal{T}=\{(\gvect_i,\xvect_i)\}_{i=1}^{I}$ and let $(\gvect,\xvect)\in\mathcal{T}$. The vector $\gvect$ is the concatenation of $K$ vectors $\gvect = [\gvect_1 | \dots \gvect_k | \dots \gvect_K]$ where $[\cdot | \cdot]$ denotes vertical vector concatenation. We recall that for all sub-bands $k ; 1\leq k \leq K$, $\gvect_k\in\mathbb{R}^{2J}$ where $J$ is the number of frequencies in each sub-band. Without loss of generality we consider the sub-band $k$.
The joint probability of $(\gvect_k,\xvect)$ can be marginalized as:
\begin{align}
\label{eq:gllim-model}
p(\gvect_k,\xvect) = \sum_{r=1}^{R} p(\gvect_k | \xvect, C_k=r) 
p(\xvect_k | C_k=r) p(C_k=r).
\end{align}
Assuming Gaussian variables, we have $p(\gvect_k | \xvect, C_k=r) = \mathcal{N}(\gvect_k | h_{kr}(\xvect) , \Sigmamat_{kr})$, $p(\xvect | C_k=r) = \mathcal{N} (\xvect | \nuvect_{kr}, \Omegamat_{kr})$, and $p(C_k=r)=\pi_{kr}$, where $h_{kr}(\xvect) = \Lmat_{kr} \xvect + \lvect_{kr}$ with
$\Lmat_{kr} \in \mathbb{R}^{2J\times 2}$ and $\lvect_{kr}\in \mathbb{R}^{2J}$, $\Sigmamat_{r} \in \mathbb{R}^{2J\times 2J}$ is the associated covariance matrix, and $\xvect$ is drawn from a Gaussian mixture model with $R$ components, each component $r$ being characterized by a prior $\pi_{kr}$, a mean $\nuvect_{kr}\in\mathbb{R}^2$ and a covariance $\Omegamat_{kr}\in\mathbb{R}^{2\times 2}$. The parameter set of this model for sub-band $k$ is:
\begin{equation}
\label{eq:parameters-theta}
\Thetavect_{k} = \{  \Lmat_{kr}, \lvect_{kr}, \Sigmamat_{kr}, \nuvect_{kr}, \Omegamat_{kr}, \pi_{kr}\}_{r=1}^{r=R}.
\end{equation}
These parameters can be estimated via a closed-form EM procedure from a training dataset, e.g. $\mathcal{T}$
(please consult 
[15], [29] and Section~VII-C). 

One should notice that there is a parameter set for each sub-band $k$, $1\leq k \leq K$, hence there are $K$ models that need be trained in our case.
It follows that 
(12) writes:
\begin{align}
\label{eq:linear-mapping}
p (  \gvect_{tk} |&  \xvect_{tn},  B_{tk} =n , C_{tk} = r)  =  \\
&
\begin{cases}
\mathcal{N} (\gvect_{tk} ; \Lmat_{kr} \xvect_{tn} + \lvect_{kr}, \Sigmamat_{kr}) & \textrm{if} \: 1\leq n \leq N \\
\mathcal{U} (\gvect_{tk} ; \textrm{vol} (\mathcal{G})) & \textrm{if} \: n=0.
\end{cases} \nonumber
\end{align}
The right-hand side of 
(7) can now be written as:
\begin{align}
\label{eq:region-posterior}
p(C_{tk}=r|\xvect_{tn}, B_{tk}=n)
 = \frac{\pi_r \mathcal{N} ( \xvect_{tn} ; \nuvect_{r}, \Omegamat_{r})}{\sum_{i=1}^{R} \pi_i \mathcal{N} ( \xvect_{tn} ; \nuvect_{i}, \Omegamat_{i})}.
\end{align}

\section{Derivation of the E-S Variational Step}
\label{app:derivations}
 The E-S step for the per-person variational posterior distribution of the state vector $q(\svect_{tn})$ is evaluated by expanding~(16), namely:
\begin{align}
 \log & \ q(\svect_{tn})  = 
    \mathbb{E}_{q(\zvect_{t}) \prod_{\ell \neq n} 
q(\svect_{t\ell})} [\log p(\fmat_t |  \smat_t,\avect_t) p(\gmat_t |  \smat_t,\bvect_t, \cvect_t) \nonumber \\
\times & p(\zvect_t | \smat_t)
 \int p(\smat_t | \smat_{t-1}) q( \smat_{t-1} | \omat_{1:t-1}) d\smat_{t-1}]   \nonumber \\
= & \sum_{m=1}^{M_t} \mathbb{E}_{q(\Avect_{t})}
 [\delta_{(\Avect_{tm}=n)} \log \mathcal{N} (\vvect_{tm}; \Pmat_{f}\svect_{tn}, \Phimat_{tm})] \nonumber\\
  + & \sum_{k=1}^{K_t} \mathbb{E}_{q(\Bvect_{t},\Cvect_{t})}
  [\delta_{(\Bvect_{tk} = n,\Cvect_{tk}=r)} \log \mathcal{N} (\gvect_{tk} ; h_{kr} (\xvect_{tn}), \Sigmamat_{kr})] \nonumber \\
 + & \ \mathbb{E}_{\prod_{\ell \neq n} q(\svect_{t\ell})} [\sum_{n=1}^{N} \log \mathcal{N}(\svect_{tn};\Dmat \svect_{t-1n},\Dmat \Gamma_{t-1n} \Dmat ^\top + \Lambda_{tn})] \nonumber
 \end{align}
where constant terms are omitted.
Using~(20) and after some algebraic derivations one obtains that $q(\svect_{tn})$ follows a Gaussian distribution, i.e.~(22), where the covariance matrix and mean vector are given by (23) and (24), respectively.

\balance


\bibliographystyle{IEEEtran}

\begin{thebibliography}{10}
\providecommand{\url}[1]{#1}
\csname url@samestyle\endcsname
\providecommand{\newblock}{\relax}
\providecommand{\bibinfo}[2]{#2}
\providecommand{\BIBentrySTDinterwordspacing}{\spaceskip=0pt\relax}
\providecommand{\BIBentryALTinterwordstretchfactor}{4}
\providecommand{\BIBentryALTinterwordspacing}{\spaceskip=\fontdimen2\font plus
\BIBentryALTinterwordstretchfactor\fontdimen3\font minus
  \fontdimen4\font\relax}
\providecommand{\BIBforeignlanguage}[2]{{%
\expandafter\ifx\csname l@#1\endcsname\relax
\typeout{** WARNING: IEEEtran.bst: No hyphenation pattern has been}%
\typeout{** loaded for the language `#1'. Using the pattern for}%
\typeout{** the default language instead.}%
\else
\language=\csname l@#1\endcsname
\fi
#2}}
\providecommand{\BIBdecl}{\relax}
\BIBdecl

\bibitem{gatica2007audiovisual}
D.~Gatica-Perez, G.~Lathoud, J.-M. Odobez, and I.~McCowan, ``Audiovisual
  probabilistic tracking of multiple speakers in meetings,'' \emph{IEEE
  Transactions on Audio, Speech and Language Processing}, vol.~15, no.~2, pp.
  601--616, 2007.

\bibitem{hospedales08structure}
T.~Hospedales and S.~Vijayakumar, ``Structure inference for {B}ayesian
  multisensory scene understanding,'' \emph{IEEE Transactions on Pattern
  Analysis and Machine Intelligence}, vol.~30, no.~12, pp. 2140--2157, 2008.

\bibitem{NavqiMiaoChambers2010}
S.~Naqvi, M.~Yu, and J.~Chambers, ``A multimodal approach to blind source
  separation of moving sources,'' \emph{IEEE Journal of Selected Topics in
  Signal Processing}, vol.~4, no.~5, pp. 895 --910, 2010.

\bibitem{kilicc2015audio}
V.~K{\i}l{\i}{\c{c}}, M.~Barnard, W.~Wang, and J.~Kittler, ``Audio assisted
  robust visual tracking with adaptive particle filtering,'' \emph{IEEE
  Transactions on Multimedia}, vol.~17, no.~2, pp. 186--200, 2015.

\bibitem{schult2015information}
N.~Schult, T.~Reineking, T.~Kluss, and C.~Zetzsche, ``Information-driven active
  audio-visual source localization,'' \emph{PloS one}, vol.~10, no.~9, 2015.

\bibitem{barnard2016mean}
M.~Barnard, W.~Wang, A.~Hilton, J.~Kittler \emph{et~al.}, ``Mean-shift and
  sparse sampling-based {SMC-PHD} filtering for audio informed visual speaker
  tracking,'' \emph{IEEE Transactions on Multimedia}, vol.~18, no.~12, pp.
  2417--2431, 2016.

\bibitem{kilicc2016mean}
V.~K{\i}l{\i}{\c{c}}, M.~Barnard, W.~Wang, A.~Hilton, and J.~Kittler,
  ``Mean-shift and sparse sampling-based {SMC-PHD} filtering for audio informed
  visual speaker tracking,'' \emph{IEEE Transactions on Multimedia}, vol.~18,
  no.~12, pp. 2417--2431, 2016.

\bibitem{ba2016line}
S.~Ba, X.~Alameda-Pineda, A.~Xompero, and R.~Horaud, ``An on-line variational
  {Bayesian} model for multi-person tracking from cluttered scenes,''
  \emph{Computer Vision and Image Understanding}, vol. 153, pp. 64--76, 2016.

\bibitem{BaeYoon2018}
S.~Bae and K.~Yoon, ``Confidence-based data association and discriminative deep
  appearance learning for robust online multi-object tracking,'' \emph{IEEE
  Transactions on Pattern Analysis and Machine Intelligence}, vol.~40, no.~3,
  pp. 595--610, March 2018.

\bibitem{valin2007robust}
J.-M. Valin, F.~Michaud, and J.~Rouat, ``Robust localization and tracking of
  simultaneous moving sound sources using beamforming and particle filtering,''
  \emph{Robotics and Autonomous Systems}, vol.~55, no.~3, pp. 216--228, 2007.

\bibitem{lombard2011tdoa}
A.~Lombard, Y.~Zheng, H.~Buchner, and W.~Kellermann, ``{TDOA} estimation for
  multiple sound sources in noisy and reverberant environments using broadband
  independent component analysis,'' \emph{IEEE Transactions on Audio, Speech,
  and Language Processing}, vol.~19, no.~6, pp. 1490--1503, 2011.

\bibitem{alameda2014geometric}
X.~Alameda-Pineda and R.~Horaud, ``A geometric approach to sound source
  localization from time-delay estimates,'' \emph{IEEE/ACM Transactions on
  Audio, Speech, and Language Processing}, vol.~22, no.~6, pp. 1082--1095,
  2014.

\bibitem{dorfan2015tree}
Y.~Dorfan and S.~Gannot, ``Tree-based recursive expectation-maximization
  algorithm for localization of acoustic sources,'' \emph{IEEE/ACM Transactions
  on Audio, Speech and Language Processing (TASLP)}, vol.~23, no.~10, pp.
  1692--1703, 2015.

\bibitem{li2016estimation}
X.~Li, L.~Girin, R.~Horaud, and S.~Gannot, ``Estimation of the direct-path
  relative transfer function for supervised sound-source localization,''
  \emph{IEEE/ACM Transactions on Audio, Speech, and Language Processing},
  vol.~24, no.~11, pp. 2171--2186, 2016.

\bibitem{li2017multiple}
X.~Li, L.~Girin, R.~Horaud, S.~Gannot, X.~Li, L.~Girin, R.~Horaud, and
  S.~Gannot, ``Multiple-speaker localization based on direct-path features and
  likelihood maximization with spatial sparsity regularization,''
  \emph{IEEE/ACM Transactions on Audio, Speech and Language Processing},
  vol.~25, no.~10, pp. 1997--2012, 2017.

\bibitem{deleforge2015colocalization}
A.~Deleforge, R.~Horaud, Y.~Y. Schechner, and L.~Girin, ``Co-localization of
  audio sources in images using binaural features and locally-linear
  regression,'' \emph{IEEE/ACM Transactions on Audio, Speech and Language
  Processing}, vol.~23, no.~4, pp. 718--731, 2015.

\bibitem{gold2011speech}
B.~Gold, N.~Morgan, and D.~Ellis, \emph{Speech and audio signal processing:
  processing and perception of speech and music}.\hskip 1em plus 0.5em minus
  0.4em\relax John Wiley \& Sons, 2011.

\bibitem{lathoud2004av16}
G.~Lathoud, J.-M. Odobez, and D.~Gatica-Perez, ``{AV16.3: An audio-visual
  corpus for speaker localization and tracking},'' in \emph{Machine Learning
  for Multimodal Interaction}.\hskip 1em plus 0.5em minus 0.4em\relax Springer,
  2004, pp. 182--195.

\bibitem{gebru2018a}
I.~D. Gebru, S.~Ba, X.~Li, and R.~Horaud, ``Audio-visual speaker diarization
  based on spatiotemporal {Bayesian} fusion,'' \emph{IEEE Transactions on
  Pattern Analysis and Machine Intelligence}, vol.~40, no.~5, pp. 1086--1099,
  2018.

\bibitem{ristic2011metric}
B.~Ristic, B.-N. Vo, D.~Clark, and B.-T. Vo, ``A metric for performance
  evaluation of multi-target tracking algorithms,'' \emph{IEEE Transactions on
  Signal Processing}, vol.~59, no.~7, pp. 3452--3457, 2011.

\bibitem{vijayasenan2012diartk}
D.~Vijayasenan and F.~Valente, ``{DiarTk: an open source toolkit for research
  in multistream speaker diarization and its application to meeting
  recordings},'' in \emph{INTERSPEECH}, Portland, OR, USA, 2012, pp.
  2170--2173.

\bibitem{checka04multiple}
N.~Checka, K.~Wilson, M.~Siracusa, and T.~Darrell, ``Multiple person and
  speaker activity tracking with a particle filter,'' in \emph{IEEE
  International Conference on Acoustics, Speech, and Signal Processing}, 2004,
  pp. 881--884.

\bibitem{liu2017particle}
Y.~Liu, W.~Wang, J.~Chambers, V.~Kilic, and A.~Hilton, ``Particle flow
  {SMC}-{PHD} filter for audio-visual multi-speaker tracking,'' in
  \emph{International Conference on Latent Variable Analysis and Signal
  Separation}, 2017, pp. 344--353.

\bibitem{liu2018non}
Y.~Liu, A.~Hilton, J.~Chambers, Y.~Zhao, and W.~Wang, ``Non-zero diffusion
  particle flow {SMC-PHD} filter for audio-visual multi-speaker tracking,''
  \emph{IEEE International Conference on Acoustics, Speech, and Signal
  Processing}, pp. 4304--4308, April 2018.

\bibitem{qian20173d}
X.~Qian, A.~Brutti, M.~Omologo, and A.~Cavallaro, ``3{D} audio-visual speaker
  tracking with an adaptive particle filter,'' in \emph{IEEE International
  Conference on Acoustics, Speech and Signal Processing}, New-Orleans,
  Louisiana, 2017, pp. 2896--2900.

\bibitem{gebru2016em}
I.~D. {Gebru}, X.~{Alameda-Pineda}, F.~{Forbes}, and R.~{Horaud}, ``{EM}
  algorithms for weighted-data clustering with application to audio-visual
  scene analysis,'' \emph{IEEE Transactions on Pattern Analysis and Machine
  Intelligence}, vol.~38, no.~12, pp. 2402--2415, 2016.

\bibitem{li:hal-01851985}
X.~Li, Y.~Ban, L.~Girin, X.~Alameda-Pineda, and R.~Horaud, ``{Online
  Localization and Tracking of Multiple Moving Speakers in Reverberant
  Environments},'' \emph{{IEEE Journal of Selected Topics in Signal
  Processing}}, vol.~13, no.~1, pp. 88--103, Mar. 2019.

\bibitem{ban:hal-01969050}
Y.~Ban, X.~Alameda-Pineda, C.~Evers, and R.~Horaud, ``{Tracking Multiple Audio
  Sources with the Von Mises Distribution and Variational EM},'' \emph{{IEEE
  Signal Processing Letters}}, vol.~26, no.~6, pp. 798 -- 802, Jun. 2019, paper
  submitted to IEEE Signal Processing Letters.

\bibitem{ban2017exploiting}
Y.~Ban, L.~Girin, X.~Alameda-Pineda, and R.~Horaud, ``Exploiting the
  complementarity of audio and visual data in multi-speaker tracking,'' in
  \emph{{IEEE ICCV Workshop on Computer Vision for Audio-Visual Media}},
  Venezia, Italy, Oct. 2017, pp. 446--454.

\bibitem{ban2018accounting}
Y.~Ban, X.~Li, X.~Alameda-Pineda, L.~Girin, and R.~Horaud, ``Accounting for
  room acoustics in audio-visual multi-speaker tracking,'' in \emph{{IEEE
  International Conference on Acoustics, Speech and Signal Processing}},
  Calgary, Alberta, Canada, Apr. 2018, pp. 6553--6557.

\bibitem{bhattacharyya1943measure}
A.~Bhattacharyya, ``On a measure of divergence between two statistical
  populations defined by their probability distributions,'' \emph{Bull.
  Calcutta Math. Soc.}, vol.~35, pp. 99--109, 1943.

\bibitem{DeleforgeForbesHoraud2015}
A.~{Deleforge}, F.~{Forbes}, and R.~{Horaud}, ``High-dimensional regression
  with {Gaussian} mixtures and partially-latent response variables,''
  \emph{Statistics and Computing}, vol.~25, no.~5, pp. 893--911, 2015.

\bibitem{bishop06pa}
C.~Bishop, \emph{Pattern Recognition and Machine Learning}.\hskip 1em plus
  0.5em minus 0.4em\relax Springer, 2006.

\bibitem{smidl06var}
V.~Smidl and A.~Quinn, \emph{The Variational {B}ayes Method in Signal
  Processing}.\hskip 1em plus 0.5em minus 0.4em\relax Springer, 2006.

\bibitem{anguera2012speaker}
X.~Anguera~Miro, S.~Bozonnet, N.~Evans, C.~Fredouille, G.~Friedland, and
  O.~Vinyals, ``Speaker diarization: A review of recent research,'' \emph{IEEE
  Transactions on Audio, Speech, and Language Processing}, vol.~20, no.~2, pp.
  356--370, 2012.

\bibitem{noulas2012multimodal}
A.~Noulas, G.~Englebienne, and B.~J.~A. Krose, ``Multimodal speaker
  diarization,'' \emph{IEEE Transactions on Pattern Analysis and Machine
  Intelligence}, vol.~34, no.~1, pp. 79--93, 2012.

\bibitem{lathoud2005sector}
G.~Lathoud and M.~Magimai-Doss, ``A sector-based, frequency-domain approach to
  detection and localization of multiple speakers,'' in \emph{IEEE
  International Conference on Acoustics, Speech, and Signal Processing},
  vol.~3.\hskip 1em plus 0.5em minus 0.4em\relax IEEE, 2005, pp. 265--268.

\bibitem{cao2017realtime}
Z.~Cao, T.~Simon, S.-E. Wei, and Y.~Sheikh, ``Realtime multi-person 2{D} pose
  estimation using part affinity fields,'' in \emph{IEEE Conference on Computer
  Vision and Pattern Recognition}, Hawaii, USA, 2017, pp. 7291--7299.

\bibitem{zheng2017person}
L.~Zheng, H.~Zhang, S.~Sun, M.~Chandraker, Y.~Yang, and Q.~Tian, ``Person
  re-identification in the wild,'' in \emph{IEEE Conference on Computer Vision
  and Pattern Recognition}, Hawaii, USA, 2017, pp. 1367--1376.

\bibitem{milan2016mot16}
A.~Milan, L.~Leal-Taix{\'e}, I.~Reid, S.~Roth, and K.~Schindler, ``Mot16: A
  benchmark for multi-object tracking,'' \emph{arXiv preprint
  arXiv:1603.00831}, 2016.

\bibitem{minotto2015multimodal}
V.~P. Minotto, C.~R. Jung, and B.~Lee, ``Multimodal multi-channel on-line
  speaker diarization using sensor fusion through {SVM},'' \emph{IEEE
  Transactions on Multimedia}, vol.~17, no.~10, pp. 1694--1705, 2015.

\end{thebibliography}

\end{document}